%% file: main.tex
\documentclass{article}

\usepackage{microtype}
\usepackage{graphicx}
\usepackage{subcaption}
\usepackage{booktabs} 

\usepackage{hyperref}

\usepackage{mathrsfs}
\usepackage[space]{grffile}
\usepackage{url}
\usepackage{lipsum}
\usepackage{tikz}
\usetikzlibrary{arrows.meta, decorations.markings, calc, fadings, shadings}
\usepackage{algorithm,algpseudocode}
\usepackage{multirow}

\usepackage[preprint]{icml2026}

\usepackage{amsmath}
\usepackage{amssymb}
\usepackage{mathtools}
\usepackage{amsthm}

\usepackage[capitalize,noabbrev]{cleveref}

\theoremstyle{plain}

\theoremstyle{definition}

\theoremstyle{remark}

\usepackage[textsize=tiny]{todonotes}

\icmltitlerunning{CODA: Coordination via On-Policy Diffusion for Multi-Agent Offline Reinforcement Learning}

\begin{document}

\twocolumn[
  \icmltitle{CODA: Coordination via On-Policy Diffusion for Multi-Agent Offline Reinforcement Learning}



  \icmlsetsymbol{equal}{*}

  \begin{icmlauthorlist}
    \icmlauthor{Marcel Hedman}{Inephany,Oxford}
    \icmlauthor{Kale-ab Abebe Tessera}{edi}
    \icmlauthor{Juan Claude Formanek}{cape}
    \icmlauthor{Anya Sims}{Oxford}
    \icmlauthor{Riccardo Zamboni}{milan}
    \icmlauthor{Trevor McInroe}{edi}
    \icmlauthor{John Torr}{Inephany}
    \icmlauthor{Elliot Fosong}{Inephany}
  \end{icmlauthorlist}

  \icmlaffiliation{Inephany}{Inephany Ltd}
  \icmlaffiliation{Oxford}{Department of Statistics, University of Oxford}
  \icmlaffiliation{cape}{University of Cape Town
Cape Town, South Africa}
\icmlaffiliation{milan}{Politecnico di Milano, Milan, Italy}
\icmlaffiliation{edi}{The University of Edinburgh}

  \icmlcorrespondingauthor{Marcel Hedman}{marcel.hedman@jesus.ox.ac.uk}

  \icmlkeywords{Multi-Agent Reinforcement Learning, Offline Reinforcement Learning, Diffusion Models, Multi-Agent Cooperative Coordination}

  \vskip 0.3in
]



\printAffiliationsAndNotice{}  

\begin{abstract}
Offline multi-agent reinforcement learning (MARL) enables policy learning from fixed datasets, but is prone to \textbf{coordination failure}: agents trained on static, off-policy data converge to suboptimal joint behaviours because they cannot co-adapt as their policies change. We introduce \textbf{CODA} (\textit{\textbf{C}oordination via \textbf{O}n-Policy \textbf{D}iffusion for Multi-\textbf{A}gent Reinforcement Learning}), a diffusion-based multi-agent trajectory generator for data augmentation that samples conditioned on the current joint policy, producing synthetic experience which reflects the evolving behaviours of the agents, thereby providing a mechanism for co-adaptation. We find that previous diffusion-based augmentation approaches are insufficient for fostering multi-agent coordination because they produce static augmented datasets that do not evolve as the current joint policy changes during training; CODA resolves this by more closely simulating on-policy learning and is a meaningful step toward coordinated behaviours in the offline setting. CODA is algorithm-agnostic and can be layered onto both model-free and model-based offline reinforcement learning pipelines as an augmentation module. Empirically, CODA not only resolves canonical coordination pathologies in continuous polynomial games but also delivers strong results on the more complex MaMuJoCo continuous-control benchmarks. 
\end{abstract}

\input{1_intro}

\input{2_background}
\input{3_method}

\input{5_experiments}
\input{4_relwork}
\input{6_discussion}

\section*{Acknowledgements}

MH is supported by funding provided by Inephany Ltd alongside Novo Nordisk
and by the EPSRC Centre for Doctoral Training in Modern
Statistics and Statistical Machine Learning (EP/S023151/1). AS is supported by the EPSRC Centre for Doctoral Training in Modern Statistics and Statistical Machine Learning (EP/S023151/1). 

\section*{Impact Statement}

This paper presents work whose goal is to advance the field of Machine
Learning. There are many potential societal consequences of our work, none
which we feel must be specifically highlighted here.

\bibliography{example_paper}
\bibliographystyle{icml2026}

\newpage
\appendix
\onecolumn
\input{7_appendix}

\end{document}

%% file: 1_intro.tex
\section{Introduction}
\label{sec:introduction}
Multi-agent reinforcement learning (MARL) has emerged as a key framework for training intelligent agents that must operate in shared environments~\citep{marl-book,maddpg,mamujoco}. From coordinated robot teams~\citep{multi-robot-exploration} to large-scale network control and economic resource allocation~\citep{economic_marl}, MARL has been used in applications in which learning optimal \emph{joint} behaviour is critical. 

While online MARL remains the dominant research paradigm, it requires interactive exploration, which may be unsafe, costly, or infeasible in real-world systems. Offline reinforcement learning (RL), instead learns from fixed datasets, promising improved safety and sample efficiency~\citep{Levine2020OfflineRL}. However, shifting from single-agent to cooperative multi-agent settings introduces additional complexities that compound the standard offline issues of distributional shift and extrapolation error~\citep{Fu2020D4RLDF}. Specifically, it requires resolving per-agent credit assignment~\citep{marl-book} and coordination demands absent in single-agent environments~\citep{tilbury2024coordinationfailurecooperativeoffline-BRUD}.

In this work, we focus on tackling \textbf{offline coordination failure} in cooperative tasks: where distinct agents converge to incompatible behaviours despite individually taking optimal actions under the pre-collected dataset~\citep{MOMAPPO, tilbury2024coordinationfailurecooperativeoffline-BRUD}. During online training, an agent’s local optimization not only shapes its own behaviour, but also alters the experiences of its teammates, who can in turn adapt. By contrast, in \emph{offline} MARL, the dataset is static, so cross-agent adaptation cannot occur even with sufficient marginal coverage of the joint state-action space in the dataset. 

A standard approach to mitigating coordination challenges online is Centralized Training with Decentralized Execution (CTDE), which uses global information during training while keeping decentralized policies at test time. CTDE underpins successful online MARL algorithms, including MAPPO~\citep{yu2022the}, MADDPG~\citep{maddpg} and MASAC~\citep{SAC}. However, recent work~\citep{MOMAPPO, tilbury2024coordinationfailurecooperativeoffline-BRUD} shows that naively porting these model-free CTDE methods to offline MARL does not prevent coordination failures. In particular, so-called Best Response under Data (BRUD) algorithms \citep{tilbury2024coordinationfailurecooperativeoffline-BRUD}, which include MADDPG and MASAC, independently optimize each agent against the fixed dataset, and can actively induce miscoordination even with centralized critics. The issue is not missing centralized information but missing \emph{co-adaptation}. Restoring this adaptation signal, without additional environment interaction, is the challenge we address.

We propose \textbf{CODA (Coordination via On-Policy Diffusion for
Multi-Agent Offline Reinforcement Learning)}, a trajectory-level augmentation method for offline MARL. We argue that offline coordination failure arises fundamentally from the absence of \emph{endogenous joint policy evolution} with respect to the dataset. In online MARL, each agent’s update alters its teammates’ effective learning environment, producing a feedback-driven drift in the joint trajectory distribution.

CODA approximates this missing co-adaptation in the offline setting by learning a diffusion model over joint trajectories and using \emph{on-policy guidance at sampling time} to condition synthetic rollouts on the current joint policy. Unlike approaches that rely on $Q$-values or behavioural statistics~\citep{tilbury2024coordinationfailurecooperativeoffline-BRUD, oh2024diffusionbasedepisodesaugmentationoffline-EAQ}, CODA produces policy-dependent joint trajectories that track the behaviours agents are currently learning, while remaining strictly offline.

As a data augmentation method, CODA is agnostic to the MARL algorithm used, so it can be layered onto existing model-free or model-based offline MARL approaches. By uniting trajectory-level diffusion with on-policy conditioning, CODA provides a step toward resolving coordination failures in offline MARL. Empirically, we demonstrate that CODA resolves explicit coordination pathologies in continuous polynomial games and achieves strong empirical performance on complex continuous control benchmarks.

%% file: 2_background.tex
\section{Background}
\label{sec:background}

\subsection{Multi-Agent Reinforcement Learning}
\label{subsec:multi_agent_reinforcement_learning}

We consider a fully cooperative multi-agent reinforcement learning (MARL) setting modelled as a Decentralized Partially Observable Markov Decision Process~\citep[Dec-POMDP,][]{POMDPs}. A Dec-POMDP is defined by the tuple,  $G=\langle I,S,\{A^i\},T,R,\{O^i\},\Omega,\gamma\rangle$,
with agents $I=\{1,\dots,N\}$. At time $t$, agent $i$ observes $o_t^i\in O^i$ and takes actions $a_t^i\in A^i$.
Let $\mathbf{o}_t=(o_t^i)_{i\in I}\in \mathbf{O}\coloneqq \prod_i O^i$ and $\mathbf{a}_t=(a_t^i)_{i\in I}\in \mathbf{A}\coloneqq \prod_i A^i$ denote the joint quantities.
Dynamics follow $T(s_{t+1}\mid s_t,\mathbf{a}_t)$, observations $\Omega(\mathbf{o}_t\mid s_t)$, and shared team reward $r_t=R(s_t,\mathbf{a}_t)$.
Finally, decentralized policies factorize as $\boldsymbol{\pi}(\mathbf{a}_t\mid \mathbf{o}_t)=\prod_i \pi^i(a_t^i\mid o_t^i)$.
The induced distribution over finite-horizon trajectories $\tau=(s_0,\mathbf{o}_0,\mathbf{a}_0,r_0,\dots,s_H)$ is
\begin{equation}
    p_{\boldsymbol{\pi}}(\tau)\! =\! p_0(s_0)\!\!\prod_{t=0}^{H-1}\!\!\Omega(\mathbf{o}_t\! \mid \! s_t)\boldsymbol{\pi}(\mathbf{a}_t\! \mid\! \mathbf{o}_t)T(s_{t+1}\! \mid\! s_t,\mathbf{a}_t),\label{eq:traj_dist}
\end{equation}
where $\boldsymbol{\pi}$ is the trainable joint policy~\citep{Levine2018ReinforcementLA, Jackson2024PolicyGuidedD}. We optimize $J(\boldsymbol{\pi})=\mathbb{E}_{\tau\sim p_{\boldsymbol{\pi}}(\tau)}[\sum_{t=0}^{H-1}\gamma^t r_t]$ where $\gamma$ is the discount factor.

\paragraph{Offline MARL.}~~In the offline setting, learning is restricted to a fixed dataset
\(
\mathcal{D}_{\mathrm{off}}
\sim p_{\boldsymbol{\pi}_{\mathrm{off}}}(\tau),
\)
collected under an unknown behaviour policy
$\boldsymbol{\pi}_{\mathrm{off}}$.
No additional environment interaction is permitted during policy learning. The objective is to learn a policy $\boldsymbol{\pi}^*$ that maximizes the expected discounted return $J(\boldsymbol{\pi}^*)$.
Throughout, we adopt the centralized training with decentralized execution (CTDE) learning paradigm, wherein centralized critics may condition on global state and joint actions during training, but at execution time each agent selects actions using only its local observation $o^i$.

Beyond standard offline single-agent RL challenges (e.g. partial observability or distributional shift between $\boldsymbol{\pi}_{\text{off}}$ and $\boldsymbol{\pi}_{\text{curr}}$), offline MARL introduces a structural difficulty: agent coordination must be inferred from static off-policy data, without interactive co-adaptation among agents. 

\subsection{Diffusion Models}
In this work we seek to generate synthetic RL trajectories. We do so using diffusion models (DM) --- a class of generative model noted for their ability to sample from complex multimodal distributions. DMs generate samples by reversing a gradual noising process that corrupts data through iterative Gaussian convolutions. Here, we leverage their ability to conduct trajectory-level modelling~\citep{10.5555/3666122.3668131}. Starting from clean data (trajectories) $\tau_0 \sim p(\tau)$, a forward process produces progressively noisier variables $\tau_x$ according to a predefined noise schedule $\sigma(x)$, such that at sufficiently large noise levels \(\sigma(x)\), the corrupted distribution \(p(\tau;\sigma(x))\) approaches an isotropic Gaussian. 

We adopt the probability flow ODE formulation of~\citet{EDMKarras}, a deterministic special case of the general SDE-based framework. Given a continuous noise schedule \( \sigma(x) \), the data evolves as:
\begin{equation}
d\tau = -\dot{\sigma}(x)\sigma(x)\nabla_\tau \log p(\tau; \sigma(x))\, \mathrm{d}x,
\end{equation}
where $\nabla_\tau \log p(\tau; \sigma(x))$ is the noise-conditioned score function and $\dot{\sigma}(x)  := \frac{d\sigma(x)}{dx}$ denotes the derivative of the noise schedule with respect to $x$. Learning reduces to estimating the score function with a neural network trained on trajectories from $\mathcal{D}_{\text{off}}$. Sampling is then performed by numerically integrating the ODE from Gaussian noise back to the data space.

\subsection{Conditional Generation}
\label{subsec:conditional_generation}
As we will see in \cref{sec:restoring_joint_with_generative_modelling}, ultimately our goal is to conduct conditional generation such that we are able to generate trajectories that are policy conditioned.
Two standard guidance mechanisms to conduct such conditioning are Classifier-Free Guidance and Classifier Guidance.

\paragraph{Classifier-Free Guidance (CFG).} CFG trains a single model to predict both conditional and unconditional scores by randomly dropping the conditioning variable $y$ during training~\citep{ho2022classifierfreediffusionguidance}. At sampling time, conditional and unconditional scores are linearly combined:
\begin{equation}
\begin{aligned}
\hat{\nabla}_\tau \log p(\tau; \sigma \mid y)
&\approx (1+w)\nabla_\tau \log p(\tau; \sigma \mid y) \\
&\quad \quad - w \nabla_\tau \log p(\tau; \sigma),
\end{aligned}
\label{eq:cfg_equation}
\end{equation}
where $w \ge 0$ is a guidance scale controlling the strength of conditioning. Larger $w$ increases adherence to the condition at the cost of reduced diversity.

\paragraph{Classifier Guidance.} Alternatively, classifier guidance augments the unconditional score using gradients from a separately trained discriminative model $p(y \mid \tau; \sigma)$~\citep{classifier_guidance}. 
\begin{equation}
\begin{aligned}
\hat{\nabla}_\tau \log p(\tau; \sigma \mid y)
&\approx \nabla_\tau \log p(\tau; \sigma) \\
&\quad \quad + \lambda \nabla_\tau \log p(y \mid \tau; \sigma),
\end{aligned}
\label{eq:classfier_based_guidance_eqn}
\end{equation}
where $\lambda$ is a guidance scale, $\nabla_\tau \log p(\tau; \sigma)$ is the standard unconditional diffusion score and $\nabla_\tau \log p(y \mid \tau; \sigma)$ is provided by a classifier (or more generally, any differentiable scoring model). In reinforcement learning contexts, such a classifier may correspond to value functions~\citep{oh2024diffusionbasedepisodesaugmentationoffline-EAQ, Janner2022PlanningWD}, policy likelihoods~\citep{Jackson2024PolicyGuidedD}, or other objectives defined over trajectories~\citep{diffusionmultitask}.

\section{Offline Coordination Failure}
\label{sec:coordination_failure}

Recent work has studied the challenge of coordination in cooperative offline MARL. 
Notably,~\citet{MOMAPPO} decompose offline coordination into two components: \emph{strategy agreement}, which entails selecting among multiple compatible optima in the joint policy space, and \emph{strategy fine-tuning}, which refers to calibrating behaviours to realize the chosen strategy. 
Both components rely on agents adapting to one another during learning. This in turn requires that individual policy updates are conditioned on the \emph{current joint policy}.

\paragraph{BRUD updates in offline MARL.}
\Citet{tilbury2024coordinationfailurecooperativeoffline-BRUD} formalized the prevailing approach to offline MARL as the Best Response Under Data (BRUD) paradigm. Multi-agent actor-critic methods have an objective of the form:
\begin{equation}
J(\boldsymbol{\pi})
=
\mathbb{E}_{\mathbf{a}\sim \boldsymbol{\pi}}
\big[
Q(\mathbf{o}, \mathbf{a}) + \alpha \mathcal{R}(\boldsymbol{\pi})
\big],
\label{eq:brud_obj}
\end{equation}
where $Q$ is a centralized critic over joint observations and actions, and $\mathcal{R}$ is a policy regularization term with strength parameter $\alpha \in \mathbb{R}$. 
This formulation underlies widely used CTDE methods such as MASAC~\citep{SAC}, MAPPO \citep{yu2022the}, and MADDPG~\citep{maddpg}.

Applying this objective in the offline setting requires constructing a joint action using both the current policy and the dataset. Following~\citet{tilbury2024coordinationfailurecooperativeoffline-BRUD}, the updated agent samples its action from the current policy, while teammate actions are drawn directly from the dataset:
\begin{equation}
a^i \sim \pi^i_{\text{curr}}(\cdot \mid o^i),
\qquad
\mathbf{a}^{-i} \sim \mathcal{D}_{\text{off}}.
\label{eq:brud}
\end{equation}

Ignoring the regularization term, the resulting offline policy gradient for agent $i$, with parameters $\theta^i$, under a deterministic actor $\boldsymbol{\mu}_\theta$ becomes:
\begin{equation}
\scalebox{0.92}{$\displaystyle
\nabla_{\theta^i} J(\boldsymbol{\mu}_\theta) =\!\!\!\!\!
\mathop{\mathbb{E}}_{\substack{\,\,\\(\mathbf{o},\mathbf{a})\sim \mathcal{D}_{\text{off}}}}\!\!\!\!
\!\left[ \nabla_{\!\theta^i} \mu_{\theta^i}(o^i)
\nabla_{\!\tilde{a}^i} Q\big(\mathbf{o}, \!(\mu_{\theta^i}(o^i), \mathbf{a}^{\!\!-i})\big)
\right].
$}
\label{eq:brud_grad}
\end{equation}
Thus, the update for agent $i$ is a best response to dataset actions rather than to teammates’ current behaviours.

\paragraph{Coordination challenges with BRUD updates.} 
In online MARL, data is continually collected under the current joint policy $\boldsymbol{\pi}_{\text{curr}}$. 
Policy updates therefore change the distribution of future experience, enabling teammates to respond to one another’s behavioural changes. 
This feedback-driven shift in the joint trajectory distribution across optimization iterations constitutes \emph{endogenous joint policy evolution}. Under idealized conditions (sufficient exploration, centralization, realisable function approximation, and stable optimization), this interaction loop allows agents to repeatedly adapt to one another, enabling both strategy agreement and fine-tuning \citep{MOMAPPO}.

By contrast, offline MARL breaks this interaction loop due to learning from a fixed dataset
\[
\mathcal{D}_{\text{off}} \sim p_{\boldsymbol{\pi}_{\text{off}}}(\tau),
\]
collected under a typically unknown behaviour policy $\boldsymbol{\pi}_{\text{off}}$. 
The sampled experience is therefore drawn from the fixed distribution $p_{\boldsymbol{\pi}_{\text{off}}}(\tau)$ even as the current policy $\boldsymbol{\pi}_{\text{curr}}$ changes during training.

Consequently, each agent updates using trajectories that reflect teammates’ randomly sampled past behaviour rather than their current policies. 
Agents may therefore observe teammate actions that are inconsistent with the joint distribution induced by the evolving current policy, leading to incorrect adaptation and miscoordination.

This induces a structural distribution mismatch. Training implicitly optimizes a surrogate objective under $p_{\boldsymbol{\pi}_{\text{off}}}(\tau)$, whereas evaluation is under $p_{\boldsymbol{\pi}_{\text{curr}}}(\tau)$:
\begin{equation}
J(\boldsymbol{\pi}_{\text{curr}})
=
\mathbb{E}_{\tau \sim p_{\boldsymbol{\pi}_{\text{curr}}}}
\!\left[\sum_{t=0}^{H-1} \gamma^t r_t\right].
\label{eq:ideal_target_optimising_target}
\end{equation}
Crucially, because $p_{\boldsymbol{\pi}_{\text{curr}}}(\tau)$ does not influence the sampled data distribution during training, the endogenous joint policy evolution present in online learning is absent in the offline setting.

\noindent\textbf{Implications for coordination.}
Under BRUD updates, each agent effectively learns a best response to teammate actions recorded in the dataset. 
Since those actions are sampled from a potentially very different policy than the current policy, the agents’ policies do not necessarily co-evolve during training. 
As a result, the learning dynamics required for reliable strategy agreement and fine-tuning may not emerge.

\paragraph{Why dataset coverage is not enough.}
Increasing dataset coverage can mitigate value-function extrapolation error, but it does not alone resolve this coordination issue. 
Even if
\(
\mathrm{supp}(p_{\boldsymbol{\pi}_{\text{curr}}}(\tau))
\subseteq
\mathrm{supp}(p_{\boldsymbol{\pi}_{\text{off}}}(\tau)),
\)
gradients computed under the offline distribution may still differ from $\nabla J(\boldsymbol{\pi}_{\text{curr}})$ because teammate actions in the dataset are not drawn from their current policies.

Each agent therefore optimizes against effectively non-adapting teammates. 
Marginal distribution coverage alone cannot ensure consistent joint strategies, and \cref{app:constant_gradients_polygames} shows that in some settings offline learning can even collapse to a constant non-adaptive gradient signal.

Taken together, the offline training process fails to reflect how the current joint policy reshapes the trajectory distribution. In this work, we address this limitation through \emph{data augmentation}, generating trajectories consistent with the evolving current policy while remaining strictly offline.

%% file: 3_method.tex
\section{Restoring Joint Policy Evolution via Generative Modelling}
\label{sec:restoring_joint_with_generative_modelling}
Offline MARL in the BRUD paradigm lacks the feedback loop whereby as the joint policy $\boldsymbol{\pi}_\text{curr}$ changes, the
distribution of trajectories sampled changes accordingly. CODA approximately restores this coupling, \emph{without further environment interaction}, by defining a policy-dependent target trajectory distribution and using conditional generation to approximately sample from it.

\subsection{The Desired Target Distribution}
Ideally, training would use trajectories from the current joint policy trajectory distribution, namely
\begin{equation}
\begin{aligned}
& p_{\boldsymbol{\pi}_{\text{curr}}}(\tau)
 \;=\; \\ &
p_0(s_0)\prod_{t=0}^{H-1} \Omega(\mathbf{o}_t\mid s_t)\,\boldsymbol{\pi}_{\text{curr}}(\mathbf{a}_t\mid \mathbf{o}_t)\,T(s_{t+1}\mid s_t,\mathbf{a}_t).
\end{aligned}
\label{eq:p_pi_target}
\end{equation}

If we could sample $\tau \sim p_{\boldsymbol{\pi}_{\text{curr}}}(\tau)$ (or equivalently
evaluate expectations under it), then training would reflect how the \emph{curr} joint policy reshapes
the joint trajectory distribution, thereby restoring endogenous joint policy evolution and enabling
co-adaptation.

\subsection{Sampling From the Target Distribution}

CODA uses conditional diffusion to generate trajectories approximately from  $p_{\boldsymbol{\pi}_{\text{curr}}}(\tau)$. Operationally, diffusion sampling requires a noise-conditioned score $\nabla_{\hat{\tau}}\log p_{\boldsymbol{\pi}_{\text{curr}}}(\hat{\tau};\sigma)$.
We denote by $\hat{\tau}$ a noised trajectory at noise level $\sigma$.

Two complementary routes are used to sample from this distribution:
(i) \textbf{policy-guided sampling} via classifier guidance when the policy is not easily encoded as an explicit conditioning variable; and
(ii) \textbf{direct conditioning} via classifier-free guidance (CFG) when a compact conditioning representation of the policy is available.

\paragraph{(A) Policy-guided score-based sampling (implicit conditioning).}
We begin by leveraging the score-based derivation of Policy-Guided Diffusion~
\citep[PGD,][]{Jackson2024PolicyGuidedD}, and adapt it to the multi-agent setting.
The ideal objective in \cref{eq:ideal_target_optimising_target} requires expectations under $p_{\boldsymbol{\pi}_{\text{curr}}}(\tau)$. Given we only have access to offline data samples, trajectory-level importance sampling can be invoked. For any test function $f(\tau)$,
\begin{equation}
\begin{aligned}
    \mathbb{E}_{\tau\sim p_{\boldsymbol{\pi}_{\text{curr}}}}&\!\left[ f(\tau)\right]
\;=\;
\mathbb{E}_{\tau\sim p_{\boldsymbol{\pi}_{\text{off}}}}\!\left[\mathbf{w}(\tau)\,f(\tau)\right],
\qquad
 \\ & \mathbf{w}(\tau)\triangleq \prod_{t=0}^{H-1}\frac{\boldsymbol{\pi}_{\text{curr}}(\mathbf{a}_t\mid \mathbf{o}_t)}{\boldsymbol{\pi}_{\text{off}}(\mathbf{a}_t\mid \mathbf{o}_t)}.
 \end{aligned}
\label{eq:traj_is}
\end{equation}

Equivalently,
\begin{align}
p_{\boldsymbol{\pi}_{\text{curr}}}(\tau)
&=
p_{\boldsymbol{\pi}_{\text{off}}}(\tau)
\prod_{t=0}^{H-1}
\frac{\boldsymbol{\pi}_{\text{curr}}(\mathbf{a}_t \mid \mathbf{o}_t)}
{\boldsymbol{\pi}_{\text{off}}(\mathbf{a}_t \mid \mathbf{o}_t)}.
\label{eq:change_measure}
\end{align}

This change-of-measure identity holds provided the standard importance sampling support condition is satisfied, namely that
\(
\mathrm{supp}(p_{\boldsymbol{\pi}_{\text{curr}}}(\tau))
\subseteq
\mathrm{supp}(p_{\boldsymbol{\pi}_{\text{off}}}(\tau))
\),
so that the ratio is well-defined. In offline MARL this assumption is rarely guaranteed in practice, as the offline dataset may not provide full coverage of the possible trajectories from the on-policy distribution. Consequently, directly implementing the exact importance ratio can be unstable or undefined when the current policy places mass outside the support of the data. We address this below.

Taking logs and differentiating \cref{eq:change_measure} yields
\begin{equation}
\begin{aligned}
& \nabla_{\!\tau} \log p_{\boldsymbol{\pi}_{\text{curr}}}(\tau)
=
\nabla_{\!\tau} \log p_{\boldsymbol{\pi}_{\text{off}}}(\tau)
 \\ &+
\sum_{t=0}^{H-1}
\Big(
\nabla_{\!\tau} \log \boldsymbol{\pi}_{\text{curr}}(\mathbf{a}_t \!\mid \!\mathbf{o}_t)\!
-\!\!
\nabla_{\!\tau} \log \boldsymbol{\pi}_{\text{off}}(\mathbf{a}_t \!\mid \!\mathbf{o}_t)
\Big).
\end{aligned}
\label{eq:score_decomp}
\end{equation}
As in prior work~\citep{Jackson2024PolicyGuidedD}, in the limit $\sigma \to 0$ the noise-conditioned score
approaches the score of the clean distribution.
Approximating the noise-conditioned current score therefore gives
\begin{equation}
\begin{aligned}
& \nabla_{\!\hat{\tau}} \log p_{\boldsymbol{\pi}_{\text{curr}}}(\hat{\tau}; \sigma)
\approx
\nabla_{\!\hat{\tau}} \log p_{\boldsymbol{\pi}_{\text{off}}}
(\hat{\tau}; \sigma)
\\ & +
\sum_{t=0}^{H-1}
\Big(
\nabla_{\!\hat{\tau}} \log \boldsymbol{\pi}_{\text{curr}}(\hat{\mathbf{a}}_t \!\mid\! \hat{\mathbf{o}}_t) \!
- \!\!
\nabla_{\!\hat{\tau}} \log \boldsymbol{\pi}_{\text{off}}(\hat{\mathbf{a}}_t \!\mid \!\hat{\mathbf{o}}_t)
\Big).
\end{aligned}
\label{eq:noised_score}
\end{equation}
The diffusion model trained on $\mathcal{D}_{\text{off}}$ provides
$\nabla_{\hat{\tau}} \log p_{\boldsymbol{\pi}_{\text{off}}}(\hat{\tau}; \sigma)$.
The current-policy term is also computable since
$\boldsymbol{\pi}_{\text{curr}}$ is known and differentiable in most RL settings.
However, $\boldsymbol{\pi}_{\text{off}}$ is typically unknown, so as in prior work, we
drop the behaviour-policy term. 

Beyond $\boldsymbol{\pi}_{\text{off}}$ being unknown, dropping this term is appropriate offline: The importance ratio implicitly assumes adequate support overlap between the current and behaviour distributions. Subtracting $\nabla\log\boldsymbol{\pi}_{\text{off}}$ can be unstable where the behaviour policy has negligible density. Keeping only the current-policy term lets the diffusion prior $p_{\boldsymbol{\pi}_{\text{off}}}$ serve as a support constraint, biasing samples toward regions likely under $\boldsymbol{\pi}_{\text{curr}}$ while staying on the empirical data manifold. As noted in~\citet{Jackson2024PolicyGuidedD}, it also limits model error or out of sample challenges.

The practical guided score becomes:
\begin{equation}
\nabla_{\hat{\tau}} \log p(\hat{\tau}; \sigma)
+
\lambda
\sum_{t=0}^{H-1}
\nabla_{\hat{\mathbf{a}}_t}
\log
\boldsymbol{\pi}_{\text{curr}}(\hat{\mathbf{a}}_t \mid \hat{\mathbf{o}}_t),
\label{eq:coda_guided_score}
\end{equation}
where $\lambda \ge 0$ is the guidance scale and gradients are
taken only with respect to action components for stability \citep{Jackson2024PolicyGuidedD}. This score corresponds to sampling from the following \emph{policy-tilted} distribution:
\begin{equation}
\tilde{p}_{\boldsymbol{\pi}_{\text{curr}}}(\tau)
\;\propto\;
p_{\boldsymbol{\pi}_{\text{off}}}(\tau)
\exp\!\Big(
\lambda
\sum_{t=0}^{H-1}
\log
\boldsymbol{\pi}_{\text{curr}}(\mathbf{a}_t \mid \mathbf{o}_t)
\Big).
\label{eq:pseudo_dist}
\end{equation}

Thus, rather than recovering $p_{\boldsymbol{\pi}_{\text{curr}}}$ exactly, we sample from a \emph{regularized surrogate} that reweights the offline trajectory distribution toward trajectories likely under the current joint policy. The diffusion prior constrains sampling to the empirical data support, while the policy term biases mass toward $\boldsymbol{\pi}_{\text{curr}}$-consistent behaviour. As $\boldsymbol{\pi}_{\text{curr}}$ evolves, this tilt updates immediately, inducing policy-dependent drift in the generated joint trajectory distribution.

\paragraph{(B) Direct policy-conditioning via CFG (explicit conditioning).} When the policy can be represented as a compact conditioning variable
(e.g., task embedding, parameter vector, or probing representation
\citep{Chandak2019LearningAR}),
we train a conditional trajectory diffusion model
$p_\theta(\tau;\sigma \mid y)$ with condition dropout and apply
classifier-free guidance (CFG) at sampling time.
This directly produces trajectories whose distribution depends on the
current joint policy descriptor $y$.

This route is preferred when a stable, low-dimensional conditioning
interface exists.

Under standard conditional diffusion training (denoising score matching with condition dropout), the learned model approximates the data conditional
\[
p_\theta(\tau \mid y)
\;\propto\;
p_{\boldsymbol{\pi}_{\text{off}}}(\tau)p(y \mid \tau).
\]
If $y$ encodes the current joint policy, then $p(y \mid \tau)$ acts as a compatibility term measuring how likely the trajectory is under that policy. In particular, when
\(
p(y \mid \tau)
\propto
\exp\!\big(
\sum_t \log \boldsymbol{\pi}_{\text{curr}}(\mathbf{a}_t \mid \mathbf{o}_t)
\big),
\)
the induced conditional coincides with the same policy-tilted distribution in
\cref{eq:pseudo_dist}. Thus, CFG likewise samples from a distribution that re-weights $p_{\boldsymbol{\pi}_{\text{off}}}(\tau)$ toward trajectories that are probable under the current joint policy.

In this sense, both score-based policy guidance and explicit conditioning via CFG target the same pseudo-distribution \cref{eq:pseudo_dist}, which serves as a tractable, support-constrained approximation to the ideal on-policy target $p_{\boldsymbol{\pi}_{\text{curr}}}(\tau)$.

\paragraph{Why this is inherently multi-agent} \Cref{eq:coda_guided_score} is joint in two critical ways: 1) The diffusion prior is trained on \emph{joint trajectories}, capturing cross-agent correlations; 2) the guidance term is the gradient of the \emph{joint} policy log-likelihood,
$
\log \boldsymbol{\pi}_{\text{curr}}(\mathbf{a}_t \mid \mathbf{o}_t)
=
\sum_{i=1}^N \log \pi^i_{\text{curr}}(a_t^i \mid o_t^i),$
so all agents’ actions are nudged simultaneously toward regions jointly consistent with their \emph{curr} decentralized policies.

\section{Implementing CODA}

We now instantiate this idea concretely. \textbf{CODA} restores endogenous joint policy evolution by (i) learning a diffusion prior over offline joint trajectories, and (ii) sampling synthetic trajectories from a policy-tilted distribution that depends on the \emph{curr} joint policy.
These trajectories can then be used to augment the offline dataset used by any offline MARL algorithm.
In \cref{Algorithms}, we provide algorithmic descriptions of the CODA data generation process (\cref{alg:coda-sampling}) and use of CODA in MARL training (\cref{alg:coda-training}).

\subsection{Full-Horizon Joint Trajectory Diffusion}

\paragraph{Object being diffused.} CODA generates finite-horizon joint trajectories $\tau=\Big(
s_0,\mathbf{o}_0,\mathbf{a}_0,r_0,\;
\dots,\;
s_H
\Big),
$ of length $H$, with $\tau \in \mathbb{R}^{d_\tau}$ after flattening and CDF normalization.
Thus, diffusion is performed over the \emph{entire horizon at once}, rather than auto-regressively.
This allows the model to capture:
(i) \textbf{temporal correlations} across time indices $t=0,\dots,H$, and 
(ii) \textbf{cross-agent correlations} through the joint quantities $\mathbf{o}_t$ and $\mathbf{a}_t$ at each time-step.

\paragraph{Centralized diffusion backbone.} We use a single centralized diffusion network that takes the noised trajectory $\hat{\tau}$ and noise level $\sigma$ and predicts the joint denoising direction. States, joint observations, joint actions, and rewards are concatenated into a unified representation, with convolutions taken over the temporal dimension. Unlike MADiff~\citep{MADIFF}, we use no explicit cross-agent attention; coordination is captured by modelling the \emph{joint} trajectory distribution over $(s_t,\mathbf{o}_t,\mathbf{a}_t,r_t)$ across the horizon. Both temporal dynamics and inter-agent dependencies are encoded directly in the learned score. We train this diffusion prior on $\mathcal{D}_{\text{off}}$ to estimate
\(
\nabla_{\hat{\tau}}\log p_{\boldsymbol{\pi}_{\text{off}}}(\hat{\tau};\sigma)
\), using the EDM framework \citep{EDMKarras}.
At sampling time, we apply the objectives from \cref{sec:restoring_joint_with_generative_modelling} to sample approximately from the target distribution \cref{eq:p_pi_target}; the two variants differ only in how conditioning is implemented.

%% file: 5_experiments.tex
\section{Experiments}
\label{sec: experiments}

We evaluate CODA across two continuous control environments: Polynomial Games \citep{tilbury2024coordinationfailurecooperativeoffline-BRUD} and MAMuJoCo \citep{mamujoco, ogmarl}. CODA is complementary to any offline MARL algorithm; a flexibility inherited from the algorithmic structure of methods such as \cite{Jackson2024PolicyGuidedD} and \citet{10.5555/3666122.3668131}. We use MADDPG~\citep{maddpg} as the underlying model-free MARL algorithm used in each baseline, in order to ensure fair comparison.

\input{5_1_polynomial_games}

\input{5_2_mamujoco}

%% file: 5_1_polynomial_games.tex
\subsection{Polynomial Games}
\label{subsec:polynomial_games}

We first consider two-player polynomial games in an offline setting, following the setup of \citet{tilbury2024coordinationfailurecooperativeoffline-BRUD}. These games provide a differentiable, continuous analogue of classic discrete matrix games \citep{MOMAPPO}. Two agents choose continuous actions $\mathbf{a}=(a^x,a^y)\in[-1,1]^2$ and receive a shared reward $R(a^x,a^y)$ given by a polynomial. The environment is single-step and stateless, so an \textit{episode} consists of a single joint action, $\mathbf{a}$, and $Q(\mathbf{a},\mathbf{o})=R(\mathbf a)$. The objective is to learn policies that select actions $\mathbf{a}$, maximizing the joint return $R(\mathbf{a})$.

\noindent\textbf{Multiplication game.}
\citet{tilbury2024coordinationfailurecooperativeoffline-BRUD} showed that BRUD algorithms such as MADDPG fail to coordinate in offline polynomial games, converging deterministically to suboptimal behaviours determined by the overall statistics in the fixed offline dataset (see \cref{app:constant_gradients_polygames} for discussion). Using MADDPG as the base offline learning algorithm, we compare CODA against three baselines: (i) using a non-augmented offline dataset; (ii) augmentation using an unconditional diffusion generator inspired by \citep{10.5555/3666122.3668131}; and (iii) augmentation using a $Q$-conditioned diffusion generator inspired by \citep{oh2024diffusionbasedepisodesaugmentationoffline-EAQ}. For this setting, we implement conditioning in CODA via CFG.

\begin{figure}[t]
    \centering
    \includegraphics[width=1\linewidth]{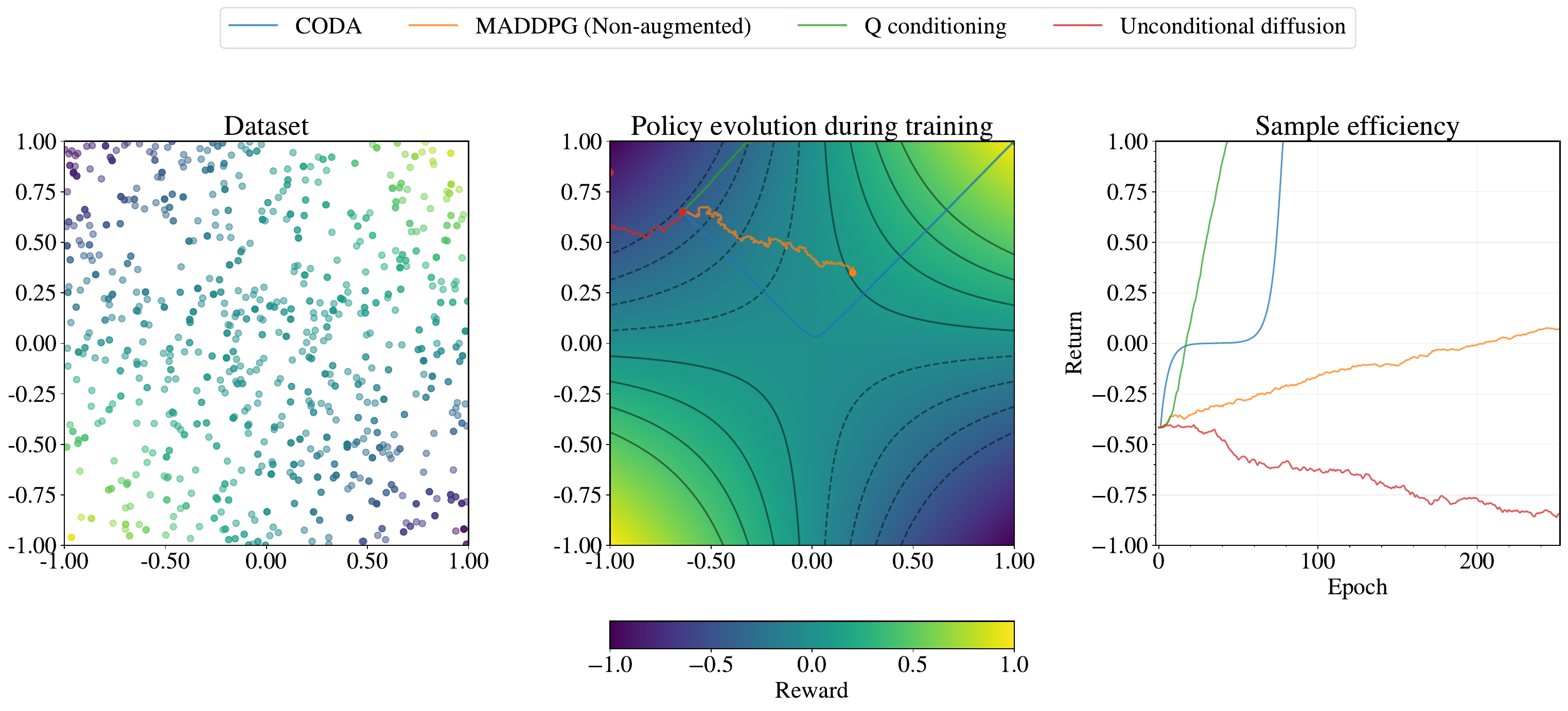}
    \caption{\textbf{Multiplication game ($R=a^xa^y$).} Reward takes values in $ [-1,1]$. Left: offline dataset used to train diffusion models and the baseline MADDPG policy. Middle: policy evolution during training. Right: returns during training (the final epoch equals the final test return for deterministic policies). Q-conditioned augmentation appears optimal only due to action-boundary effects ($\mathbf a\in[-1,1]^2$): the $y$-agent quickly saturates at the action bounds, so the deterministic gradient update acts solely through the $x$-agent, inflating returns (see \cref{fig:CODA vs baseline polynomial twin peaks game} where exploiting boundary-effects cannot reach an optimum). Sample efficiency plot contains no error bars as in this environment deterministic policies create deterministic rewards at test-time.}
    \label{fig:CODA vs baseline polynomial multiplication game}
\end{figure}

\Cref{fig:CODA vs baseline polynomial multiplication game} shows that baseline MADDPG with a non-augmented dataset converges suboptimally despite full coverage of the action space in the offline dataset, showing that coordination is not purely a coverage issue. Unconditional diffusion approximately reproduces the empirical data distribution $p_{\boldsymbol{\pi}_{\mathrm{off}}}$ and therefore inherits the same deterministic training pathology as under the static dataset. Conditioning on high $Q$ generates a high-return biased dataset but the data distribution remains static, leaving the coordination problem unresolved. The resulting training exhibits deterministic behaviour and the policy quickly collides with the action boundary during training. The convergence toward the optimum is therefore driven by the action space constraints rather than correct gradient following: The policy is dragged along the boundary. In contrast, CODA's on-policy conditioning alters the effective training distribution, allowing agents to follow the correct joint gradient and converge to the optimum.

\begin{figure}[t]
    \centering
    \includegraphics[width=1\linewidth]{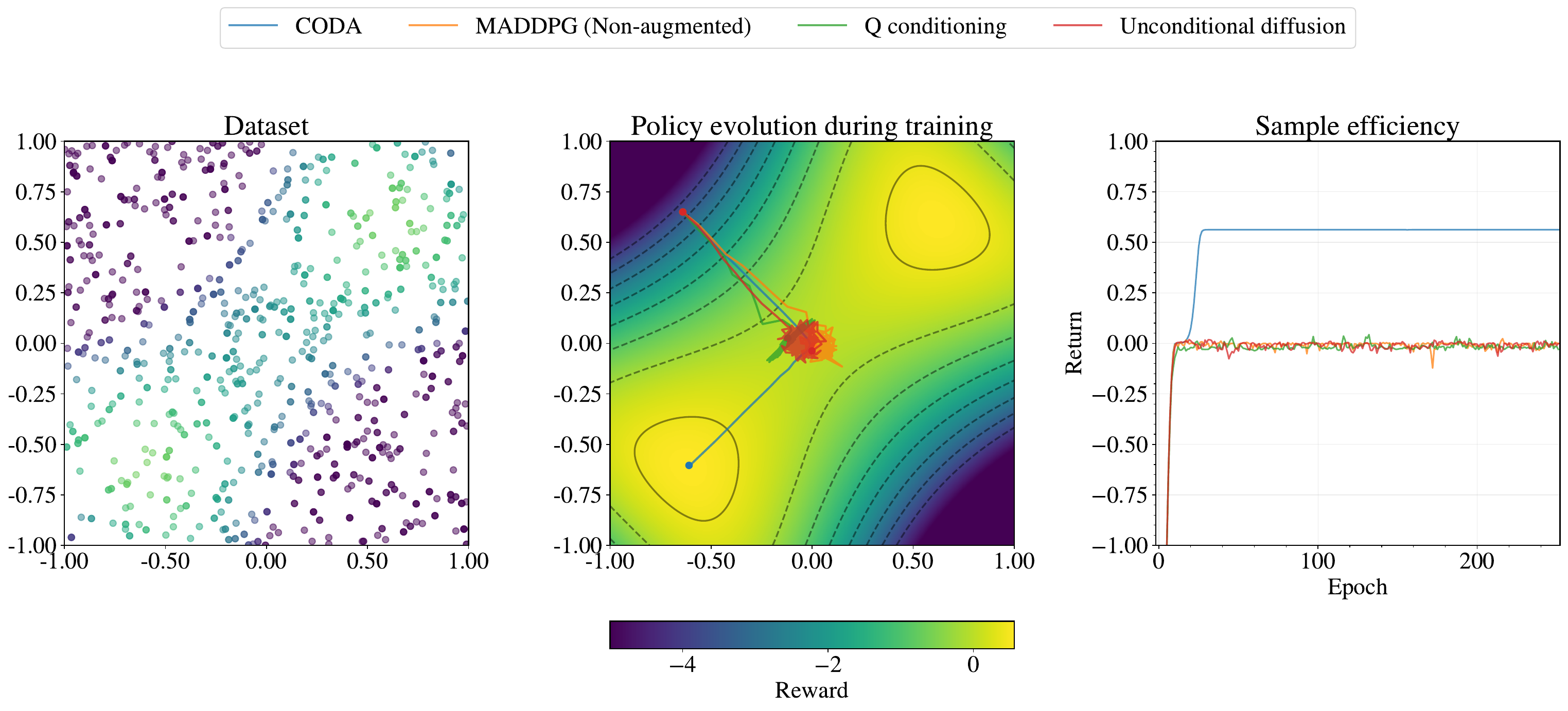}
    \caption{\textbf{Twin Peaks game.} Left: offline dataset used to train diffusion models and the baseline MADDPG policy. Middle: policy evolution during training. Right: returns during training (the final epoch equals the final test return for deterministic policies). CODA’s on-policy conditioning again mitigates offline miscoordination. Sample efficiency plot contains no error bars as in this environment deterministic policies create deterministic rewards at test-time.}
    \label{fig:CODA vs baseline polynomial twin peaks game}
\end{figure}

\paragraph{Robustness to increased agent interaction.}
To test robustness under stronger agent coupling, we consider the Twin Peaks game with reward
$R = -A((a^x)^2 + (a^y)^2) - B(a^x a^y)^2 + C a^x a^y,
\quad A>0,\;B>0,\;C>2A.$
\Cref{fig:CODA vs baseline polynomial twin peaks game} shows that CODA again recovers coordinated learning ($A{=}1,B{=}4,C{=}5$) while the baselines converge to $(0,0)$. See \cref{subsec:twin_peak_theory} for a theoretical treatment explaining the convergence to $(0,0)$. 

%% file: 5_2_mamujoco.tex
\subsection{MaMuJoCo}
\label{subsec:mamujoc_experiments}
Next, we evaluate CODA in MaMuJoCo, a higher-dimensional continuous-control benchmark in which success can depend on multiple interacting factors rather than coordination alone. We test two questions: (i) can CODA steer generated trajectories toward the target on-policy distribution; and (ii) how does this steering translate into varying degrees of downstream performance when training with a standard MARL algorithm (MADDPG) on synthetic data?

\looseness=-1
We use the BRUD-style MADDPG+BC algorithm \citep{maddpg} as the base learner and follow the MaMuJoCo setup of \citet{ogmarl}. We sweep to select the behaviour cloning coefficient. To reduce wall-clock cost, we generate synthetic trajectories periodically rather than continuously, with $N_{\textbf{epochs}}=4$ epochs between generation rounds, following the update frequency used in \citet{Jackson2024PolicyGuidedD}. Because MaMuJoCo policies are represented by high-dimensional neural networks, we use the classifier-guidance variant of CODA to target \cref{eq:pseudo_dist}.

\looseness=-1
\subsubsection{Steering trajectories on-policy} Before reporting downstream performance, we verify that the guidance scale $\lambda$ has the intended effect. For synthetic trajectories $\tau$ generated by CODA, we compute the mean action log-likelihood under the current policy $\boldsymbol{\pi}_{\text{current}}$. For deterministic policies, we evaluate this under a Gaussian surrogate centered at $\boldsymbol{\mu}_{\text{current}}(\mathbf{o}_t)$ (summed over the horizon and agents), with fixed standard deviation $1$. See \cref{app:theory_guidance_coeff} for the theoretical analysis of how $\lambda$ affects this metric.
\begin{figure}
    \centering
    \includegraphics[width=\linewidth]{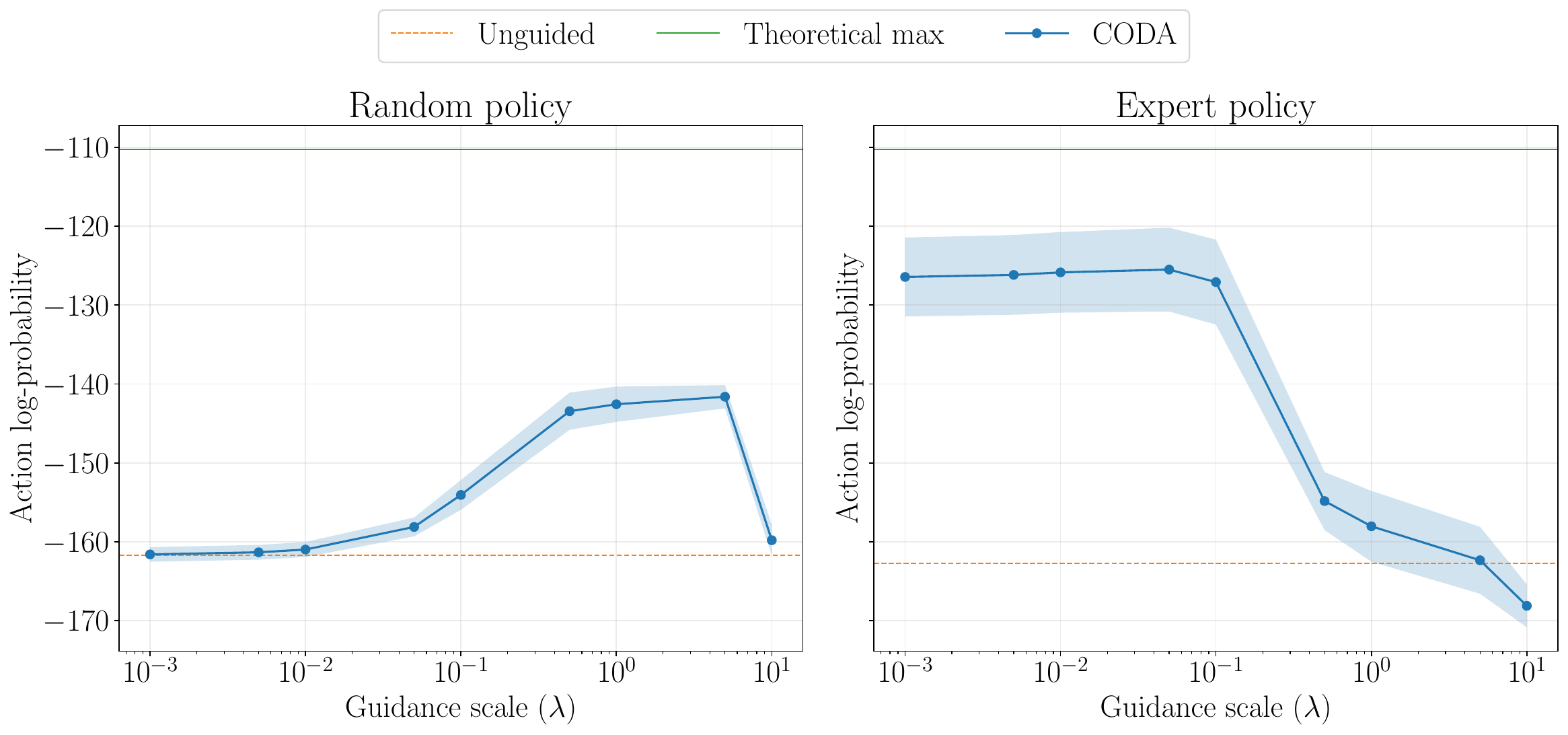}
    \caption{\textbf{On-policy steering in MaMuJoCo (2HalfCheetah).} Mean action log-likelihood of CODA-generated trajectories under the current policy $\boldsymbol \pi_{\text{current}}$ (Replay dataset). ``Random'' and ``Expert'' denote policies at initialization and after 300k training steps on the Replay dataset. Standard error across 8 seeds; horizon $H=20$; batch size 2048 trajectories. Unguided diffusion corresponds to $\lambda=0$.}
    \label{fig:log_likelihood}
\end{figure}
\looseness=-1
\paragraph{Results.}
\Cref{fig:log_likelihood} shows that unconditional diffusion ($\lambda=0$) produces trajectories with low likelihood under $\boldsymbol \pi_{\text{current}}$, while conditioning increases on-policy likelihood and exhibits the expected peak-and-collapse shape as $\lambda$ grows (See \cref{app:theory_guidance_coeff} for a theoretical explanation). Notably, for the expert target, very small $\lambda$ yields a sharp discontinuous jump in likelihood. We hypothesise this is because the expert’s action distribution is concentrated; any guidance, regardless of how small, is sufficient to move samples into a high-density region of $\boldsymbol{\pi}_{\text{current}}$.

\looseness=-1
\subsection{Investigating downstream performance}
\begin{figure}[t]
    \centering
    \includegraphics[width=1\linewidth]{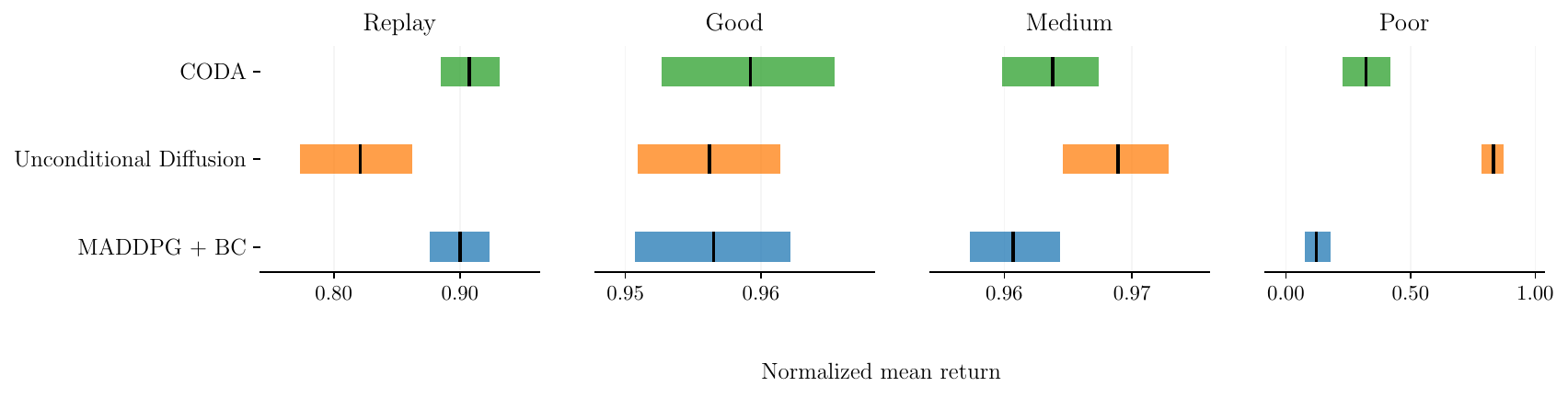}
        \caption{\textbf{2HalfCheetah mean normalized performance across datasets.} Normalized within each dataset. Standard error across 16 seeds; each seed averaged over 10 episodes.}
    \label{fig:halfcheetah-retur-norm-across-datasets}
\end{figure}

Having verified that guidance can increase likelihood under $\boldsymbol{\pi}_{\text{current}}$, we evaluate whether this improves performance when learning with MADDPG+BC using synthetic trajectories. We report test-time episodic return (mean over 10 episodes per seed) and aggregate over 16 seeds. 
\paragraph{Results.}
\Cref{fig:halfcheetah-retur-norm-across-datasets,tab:2halfcheetah} summarize final performance on 2HalfCheetah datasets and on 4Ant-Replay for (i) MADDPG+BC, (ii) MADDPG+BC with unconditional diffusion augmentation, and (iii) CODA. CODA yields the strongest gains on the \textbf{Replay} and \textbf{Good} datasets. This matches CODA’s objective: sampling from the policy-tilted distribution in \cref{eq:pseudo_dist} reweights offline data toward trajectories that remain under the support of the empirical dataset while being high-likelihood under the current policy. These datasets have examples of expert trajectories.
On \textbf{Medium} and especially \textbf{Poor}, CODA does not outperform unconditional diffusion. A plausible explanation is limited dataset support: as the learned policy improves, trajectories consistent with $\boldsymbol \pi_{\text{current}}$ may increasingly fall outside the behavioural support represented in these datasets, constraining the extent to which policy-tilting can help while remaining faithful to offline support. In contrast, unconditional diffusion can still help by providing broader transition diversity without explicitly concentrating probability mass near $\boldsymbol \pi_{\text{current}}$.
\begin{table*}
    \centering
        \caption{\textbf{MaMuJoCo performance (mean $\pm$ s.e.).} Evaluation across 16 seeds; each seed averaged over 10 episodes.}
    \begin{tabular}{c|c|cc|c}
    \toprule
         Task & Dataset & MADDPG + BC & MADDPG + BC + Diffusion  & CODA \\
         \midrule
         \multirow{4}{*}{2HalfCheetah} & Replay & \ \textbf{6854.97} $\pm$ 79.74 & 6438.85 $\pm$ 135.27 &  \textbf{6925.02} $\pm$ 72.94 \\
          & Good & \textbf{7187.77} $\pm$ 22.94 & 7185.59 $\pm$ 20.96   & \textbf{7208.19} $\pm$ 25.35 \\
         & Medium & 4515.81 $\pm$ 8.91 & \textbf{4554.11} $\pm$ 10.28 &   4530.19 $\pm$  9.41 \\
           & Poor & -64.79 $\pm$ 106.92  & \textbf{2713.69} $\pm$ 92.70 &  715.90 $\pm$ 196.56 \\
         \bottomrule
          \bottomrule
          4Ant
            & Replay & \textbf{1858.39} $\pm$ 60.19 & 1742.14
 $\pm$ 89.28 &   \textbf{1875.51} $\pm$ 80.90 \\
     \bottomrule
    \end{tabular}
    \label{tab:2halfcheetah}
\end{table*}
     

%% file: 4_relwork.tex
\section{Related Work}
\label{sec:rel_work}

\paragraph{Offline MARL and offline RL baselines.}
Offline MARL learns coordinated policies from fixed datasets, often by extending single-agent offline RL methods such as BCQ~\citep{pmlr-v97-fujimoto19a-OFF_POLICY}, CQL~\citep{kumar2020conservative}, and TD3+BC~\citep{fujimoto2021minimalist} to the decentralised MARL setting~\citep{Jiang2021OfflineDM}. To limit out-of-distribution generalisation error, ICQ~\citep{yang2021believe} imposes an implicit constraint that keeps learning near the dataset support, mitigating extrapolation error that can amplify with more agents. Complementarily, OMAR~\citep{OMAR} addresses optimisation issues from conservative value estimates by combining first-order policy gradients with a zeroth-order actor update to escape poor local optima.
\paragraph{Coordination and offline coordination failure.}
Cooperative MARL introduces per-agent credit assignment and coordination challenges absent in single-agent RL~\citep{marl-book, tilbury2024coordinationfailurecooperativeoffline-BRUD}, exacerbated by partial observability and limited information sharing~\citep{coord_MARL_limited_comms}. Recent work shows that offline cooperative MARL can fail even under CTDE because offline training is off-policy \citep{tilbury2024coordinationfailurecooperativeoffline-BRUD, MOMAPPO}. Related perspectives on coordination and generalization include plan- and learning-based coordination~\citep{plan_learn_coord}, zero-shot coordination~\citep{zero_shot_coordination}, and augmentation/self-play mechanisms~\citep{augmenting_self_play}.
\paragraph{Model-based offline MARL.}
A complementary direction to ours approximates online interaction through learned dynamics and imagined rollouts. MOMA-PPO~\citep{MOMAPPO} uses a world model to produce policy-dependent experience and reduce offline coordination failures. Beyond tackling coordination, model-based MARL learns dynamics for rollout-based updates and planning~\citep{mambpo, model_policy_marl, Zhang2021ModelbasedMP}. These approaches simulate environment transitions forward, whereas we generate joint trajectories directly from the offline data distribution with conditioning on current policies.
\paragraph{Diffusion models for RL and MARL augmentation.}
Diffusion models have been used in RL for trajectory planning~\citep{Janner2022PlanningWD, adaptdiffuser}, as expressive policy parameterizations in offline RL~\citep{Ajay2022IsCG, Wang2022DiffusionPA, Pearce2023ImitatingHB}, and as world models for imagination and long-horizon prediction~\citep{DM_world_model_atari, ding2024diffusionworldmodelfuture}. They have also been explored for single-task \citep{10.5555/3666122.3668131} \& multi-task trajectory generation~\citep{diffusionmultitask} and trust-region style control of policy updates in offline RL~\citep{DiffusionTrustRegions}. In offline MARL, diffusion-based generation has been used for episode augmentation and structured trajectory modelling: EAQ~\citep{oh2024diffusionbasedepisodesaugmentationoffline-EAQ} generates synthetic episodes via reward-conditioned diffusion to bias augmentation toward high-return trajectories, while MADiff~\citep{MADIFF} and related methods model multi-agent trajectories with diffusion backbones~\citep{yuan2025efficient, li2025dof, li2023conservatismdiffusionpoliciesoffline}. Our work builds most directly on policy-guided diffusion for single-agent RL~\citep{Jackson2024PolicyGuidedD, rigter2024a}, and extends the idea of conditioning generation on the \emph{current} policy to the multi-agent regime to tackle coordination.

%% file: 6_discussion.tex
\section{Conclusion}
This paper argues that a core driver of coordination failure in cooperative offline MARL is the loss of the feedback loop that allows teammates to adapt to one another during learning. Focusing on BRUD-style algorithms (e.g. MADDPG and MASAC), we show that existing diffusion-based augmentation approaches still yield effectively static training distributions: agents update against stale teammate behaviours and can converge to jointly suboptimal solutions, even when the offline data provides broad marginal coverage.

CODA addresses this by making augmentation policy-dependent: the generated joint trajectories are reweighted toward samples the current joint policy would most-likely produce, while remaining anchored to the empirical dataset support. In controlled polynomial games, this is sufficient to recover the desired joint optimization dynamics and avoid the deterministic miscoordination observed under BRUD-style learning and existing diffusion augmentation approaches. In MaMuJoCo, increasing guidance reliably steers generated trajectories toward the current policy and yields performance gains when dataset support is adequate, with benefits diminishing as support becomes limited.

There are a number of future work directions including adaptive guidance selection, and exploring hybrid approaches that combine policy-conditioned generation with model-based rollouts to expand effective coverage beyond the offline data while mitigating compounding model error from autoregressive world modelling.

%% file: 7_appendix.tex
\newpage
\appendix
\section{Algorithms}
\label{Algorithms}
\input{3_algorithms}

\section{Geometric Intuition for Guidance Coefficient}
\label{app:theory_guidance_coeff}
\paragraph{Convex–contractiveness of policy conditioning.}
The below investigates policy conditioning when implementing CODA using the score based classifier-based conditioning. For a deterministic policy $\mu(\mathbf o)$ with a surrogate quadratic log-likelihood model
\[
\log p(\mathbf a \mid \mathbf o ) \;=\; -\tfrac{1}{2}\|\mathbf a - \mu(\mathbf o)\|^2,
\]
the gradient of the log-likelihood is
\[
\nabla_{\mathbf a} \log p(\mathbf a \mid \mathbf o ) \;=\; - (\mathbf a - \mu(\mathbf o)).
\]
Which is a vector pointing towards $\mu(\mathbf o)$.
A single policy-conditioned update of magnitude $\lambda$ therefore takes the form
\[
\mathbf a' \;=\; \mathbf a + \lambda \nabla_{\mathbf a} \log p(\mathbf a \mid \mathbf o)
      \;=\; (1-\lambda) \mathbf a + \lambda \mu(\mathbf o).
\]
For $\lambda \in (0,2)$ this update is a \emph{convex combination} of $\mathbf a$ and $\mu(\mathbf o)$, hence strictly contractive:
\[
\|\mathbf a' - \mu(\mathbf o)\| \;=\; |1-\lambda|\|\mathbf a - \mu(\mathbf o)\|.
\]
Where the above is the distance from $\mu$ following the update step. 

\begin{itemize}
  \item If $\lambda \in (0,1)$, then $1-\lambda \in (0,1)$, so $\mathbf a'$ is a convex combination of $\mathbf a$ and $\mu(\mathbf o)$ and the distance shrinks strictly:
  \[
    \|\mathbf a' - \mu(\mathbf o)\| \;=\; (1-\lambda)\,\|\mathbf a - \mu(\mathbf o)\|
    \;<\; \|\mathbf a - \mu(\mathbf o)\|
    \qquad (\mathbf a \neq \mu(\mathbf o)).
  \]

  \item If $\lambda \in (1,2)$, then $1-\lambda < 0$, so the update overshoots to the other side of $\mu(\mathbf o)$ (the direction flips),
  but it is still contractive because $|1-\lambda| = \lambda - 1 \in (0,1)$:
  \[
    \|\mathbf a' - \mu(\mathbf o)\| \;=\; (\lambda-1)\,\|\mathbf a - \mu(\mathbf o)\|
    \;<\; \|\mathbf a - \mu(\mathbf o)\|.
  \]

  \item If $\lambda = 1$, then $\mathbf a' = \mu(\mathbf o)$ in one step.

  \item If $\lambda > 2$ (or $\lambda < 0$), then $|1-\lambda| > 1$ and the update becomes expansive:
  \[
    \|\mathbf a' - \mu(\mathbf o)\| \;>\; \|\mathbf a - \mu(\mathbf o)\|.
  \]
\end{itemize}

Thus, for $\lambda \in (0,2)$, the update always increases the Gaussian log-likelihood.  

\medskip
\noindent
In diffusion-based planners such as EDM, this idealized threshold is substantially altered by three factors.
(i) Gradients are normalized in our implementation, making the effective overshoot condition depend on the 
current distance $\|\mathbf a - \mu(\mathbf o)\|$ rather than $\lambda=1$.
(ii) Guidance is applied repeatedly across $K$ denoising steps, and later steps may partially correct earlier overshoots.
(iii) Trajectory likelihood aggregates over time and agents, so local overshoot effects are smoothed.
Consequently, we often observe that policy likelihood continues to improve even for $\lambda > 1$, 
followed by a sharp collapse once the cumulative expansive effect dominates the corrective capacity of subsequent denoising steps.

\section{BRUD Updates can Induce Constant Gradients}
\label{app:constant_gradients_polygames}

We focus this theoretical analysis within the polynomial game environment. In order to test coordination as defined by \citet{MOMAPPO}, we must test \emph{strategy agreement}.
Consequently, it is critical that the games selected admit \emph{multiple optima}: agreement is only meaningful when
there are several distinct optimal joint strategies that agents must select consistently. In contrast, in
a single-optimum setting, many methods would resolve the task trivially, since no coordination (in the form
of selecting among competing optimal conventions) is required.

\citet{tilbury2024coordinationfailurecooperativeoffline-BRUD} show that in offline MARL,
BRUD-style policy updates can induce \emph{non-adaptive} (constant) gradient fields: the
direction of improvement is determined by dataset statistics rather than by the local geometry
at the current joint policy. The below treatment is largely a restatement of their arguments.

\subsection{BRUD Updates in Stateless Polynomial Two-Player Games}
Consider the setup from \cref{subsec:polynomial_games} with continuous actions
$a=(a^x,a^y)\in[-1,1]^2$ and shared reward $R(a_x,a_y)$.
Since the environment is single-step, stateless and the reward is assumed known, the centralized critic reduces to the reward:
$Q(\mathbf o, \mathbf a)=R(a^x,a^y)$. Note a learned critic can also be implemented if desired.

We use state-independent deterministic policies $\mu_{\theta^x},\mu_{\theta^y}$ defined by
\[
\mu_{\theta^x}\equiv \theta^x,\qquad \mu_{\theta^y}\equiv \theta^y,
\]
so the chosen actions are $a^x=\mu_{\theta^x}$ and $a^y=\mu_{\theta^y}$. Then $\nabla_{\theta^x} \mu_{\theta^x}=1$, and following the action sampling approach of \cref{eq:brud}, we find that \cref{eq:brud_grad} becomes the following when updating agent $x$:
\begin{align}
\nabla_{\theta^x} J
&= \mathbb{E}_{a^y\sim \mathcal{D}_{\text{off}}}
\left[
\nabla_{\theta^x} \mu_{\theta^x}\;
\nabla_{\tilde a^x} R(\tilde a^x,a^y)\big|_{\tilde a^x= \mu_{\theta^x}}
\right]
\nonumber\\
&=
\mathbb{E}_{a^y\sim \mathcal{D}_{\text{off}}}
\left[
\partial_{\tilde a^x} R(\tilde a^x,a^y)\big|_{\tilde a^x=\theta^x}
\right].
\label{eq:brud_poly_update_x}
\end{align}
Similarly,
\begin{equation}
\nabla_{\theta^y} J
=
\mathbb{E}_{a^x\sim \mathcal{D}_{\text{off}}}
\left[
\partial_{a^y} R(a^x,a^y)\big|_{a^y=\theta^y}
\right].
\label{eq:brud_poly_update_y}
\end{equation}

Equations~\eqref{eq:brud_poly_update_x}--\eqref{eq:brud_poly_update_y} illustrate a key mismatch in offline BRUD updates.
In coordinated settings, we would like agent $x$'s update to respond to agent $y$'s \emph{current behaviour}, so that learning adapts as $y$'s policy changes.
Instead, BRUD takes an expectation over the offline dataset for the other agent's action, which marginalises out $y$'s current policy entirely.
As a result, agent $x$ does not adapt to $\mu_{\theta^y}$; any influence of agent $y$ enters only through fixed statistics of the dataset.

In the special case where the relevant term is linear in $a^y$, this dependence reduces simply to the dataset mean action of agent $y$.
More generally, for higher-order interactions it reduces to higher-order dataset moments (e.g.\ mean, variance, etc.).
In all cases, BRUD couples agents through static dataset statistics rather than through the teammate’s evolving behaviour.

\subsection{Multiplication Game: A Constant Vector Field}
\label{subsec:multiplication_gradients}
Here, we specifically consider the multiplication polynomial game $
R(a^x,a^y)=a^x a^y.$

\paragraph{Online objective (on-policy joint distribution).}
The online objective is
\begin{equation}
J_{\text{on}}(\mu_{\theta^x},\mu_{\theta^y})
\;\triangleq\;
\mathbb{E}_{a^x\sim \mu_{\theta^x},\; a^y\sim \mu_{\theta^y}}\!\left[R(a^x,a^y)\right]
=
\mathbb{E}_{a^x\sim \mu_{\theta^x}}\!\left[a^x\right]\;
\mathbb{E}_{a^y\sim \mu_{\theta^y}}\!\left[a^y\right],
\label{eq:Jon_mult}
\end{equation}
where the factorisation follows because $R(a^x,a^y)=a^x a^y$ and the sampling is independent.

\Cref{eq:Jon_mult} therefore reduces to
\begin{equation}
J_{\text{on}}(\mu_{\theta^x},\mu_{\theta^y})=\theta^x \theta^y,
\qquad\Rightarrow\qquad
\nabla_\theta J_{\text{on}}(\mu_{\theta^x},\mu_{\theta^y})=(\theta^y,\theta^x).
\label{eq:grad_online_mult}
\end{equation}
Thus the online gradient field is \emph{policy-adaptive}: the update direction varies with the
current joint policy $(\theta^x,\theta^y)$ and matches the true surface gradient
$\nabla R(a^x,a^y)=(a^y,a^x)$ evaluated at $(a^x,a^y)=(\theta^x,\theta^y)$.

\paragraph{Constant vector field under BRUD.} Under offline BRUD-style updates, using \cref{eq:brud_poly_update_x}--\cref{eq:brud_poly_update_y}, we see the below in the multiplication game setting
\[
\nabla_{\theta^x}J
= \mathbb{E}_{a^y\sim\mathcal{D}_{\text{off}}}[a^y]
\equiv \bar a^y,
\qquad
\nabla_{\theta^y}J
= \mathbb{E}_{a^x\sim\mathcal{D}_{\text{off}}}[a^x]
\equiv \bar a^x.
\]
Hence the induced BRUD ``gradient field'' in policy-parameter space is
\begin{equation}
\nabla_\theta J(\theta^x,\theta^y)
=
(\bar a^y,\bar a^x),
\label{eq:constant_field_multiplication}
\end{equation}
a \emph{constant} (unidirectional) vector determined entirely by dataset means.
In practice we sample mini-batches, so we follow an empirical estimate
$(\hat{\mathbb{E}}[a^y],\hat{\mathbb{E}}[a^x])$, which introduces small stochastic deviations
around the straight-line trajectory implied by \cref{eq:constant_field_multiplication}
(consistent with the slight wobble observed in
\cref{fig:CODA vs baseline polynomial multiplication game} by non-CODA approaches).

\paragraph{Action bounds and boundary effects.}
When actions are constrained (e.g.\ $\mathbf a\in[-1,1]^2$), a constant drift direction can push policies
quickly to the boundary; subsequent updates may then ``slide'' along the constraint surface.
This can create the appearance of improvement even when the underlying update direction is actually constant because the policy slides along the boundary.

\subsection{Twin Peaks: Dependence on Dataset Mean and Variance}
\label{subsec:twin_peak_theory}
Again following the treatment within \citet{tilbury2024coordinationfailurecooperativeoffline-BRUD}, now consider the Twin Peaks reward:
\begin{equation}
R(a^x,a^y)
=
-A((a{^x})^2+(a^y)^2)\;-\;B(a^x a^y)^2\;+\;C a^x a^y,
\qquad
A>0,\;B>0,\;C>2A.
\label{eq:twin_peaks_reward}
\end{equation}
The partial derivative for agent $x$ is
\[
\partial_{a^x} R(a^x,a^y)
=
-2A a^x\;-\;2B\,a^x\,(a^y)^2\;+\;C a^y.
\]
Plugging into \cref{eq:brud_poly_update_x} and writing dataset moments
$\bar a^y \equiv \mathbb{E}_{\mathcal{D}_{\text{off}}}[a^y]$ and
$\mathbb{E}_{\mathcal{D}_{\text{off}}}[(a^y)^2]= ( \bar a^y)^{\,2}+(\sigma^y)^2$ (where $\sigma^y$ is the
sample standard deviation of the $a^y$ in $\mathcal{D}_{\text{off}}$),
we obtain the BRUD update for $x$:
\begin{align}
\nabla_{\theta^x} J(\theta^x)
&=
\mathbb{E}_{a^y\sim{\mathcal{D}_{\text{off}}}}
\left[
-2A\theta^x - 2B\,\theta^x\,(a^y)^2 + C a^y
\right]
\nonumber\\
&=
-2A\theta^x\;-\;2B\,\theta^x((\bar a^y)^{\,2}+(\sigma^y)^2)\;+\;C\bar a^y.
\label{eq:twinpeaks_brud_grad_x}
\end{align}
By symmetry,
\begin{equation}
\nabla_{\theta^y} J(\theta^y)
=
-2A\theta^y\;-\;2B\,\theta^y((\bar a^x)^{\,2}+(\sigma^x)^2)\;+\;C\bar a^x.
\label{eq:twinpeaks_brud_grad_y}
\end{equation}

\paragraph{Fixed point (converged policy) under offline BRUD.}
Setting \cref{eq:twinpeaks_brud_grad_x} to zero yields the stationary point for agent $x$:
\begin{equation}
\nabla_{\theta^x}J=0
\quad\Longleftrightarrow\quad
\theta^{x\ast}
=
\frac{C\,(\bar a^y)}{2A + 2B((\bar a^y)^{\,2}+(\sigma^y)^2)}.
\label{eq:twinpeaks_fixed_point}
\end{equation}
This highlights that the offline BRUD solution once again does not depend on the current policy of the other agent. It depends on the dataset mean $\bar a^y$ alongside the dataset \emph{spread} $(\sigma^y)^2$ through the interaction term $(a^xa^y)^2$.

\paragraph{Origin-centred datasets cannot recover the true optimum.}
If the dataset is centred so that $\bar a^y=0$, then \cref{eq:twinpeaks_fixed_point} gives
$\theta^{x\ast}=0$ regardless of $(\sigma^y)^2$.
Thus, increasing action diversity (variance) cannot recover the true optimum under BRUD when the
dataset is centred at the origin; learning collapses to $(0,0)$ (\cref{fig:CODA vs baseline polynomial twin peaks game}).

\paragraph{Increasing diversity can be harmful even when centred near the optimum.}
Even if the dataset mean is centred at an optimal action (or near it), \cref{eq:twinpeaks_fixed_point}
shows that as $(\sigma^y)^2$ grows, the denominator increases and $\theta^{x\ast}\to 0$.
Hence, greater diversity in teammate actions within the dataset can drive the BRUD fixed point \emph{away} from the true
optimum and back toward the origin, a counter-intuitive effect induced by higher-order interaction terms. Increased action space coverage becomes detrimental.

\paragraph{How this relates to ``constant'' gradients.}
Unlike the multiplication game, Twin Peaks does not produce a strictly constant vector field. However, it is non-adaptive in relation to the current policy of teammates. \Cref{eq:twinpeaks_brud_grad_x}--\cref{eq:twinpeaks_brud_grad_y} show the same structural issue: the
\emph{cross-agent} signal entering each agent's update is mediated only through low-order dataset moments
(mean and variance), rather than through the teammate's \emph{current} action. As a result, the induced
learning dynamics becomes largely dataset-determined and only weakly responsive to the evolving joint
policy, which is the core non-adaptivity phenomenon highlighted by \citet{tilbury2024coordinationfailurecooperativeoffline-BRUD}.

\section{MaMuJoCo Additional Results}

\begin{figure}[H]
    \centering
    \includegraphics[width=1\linewidth]{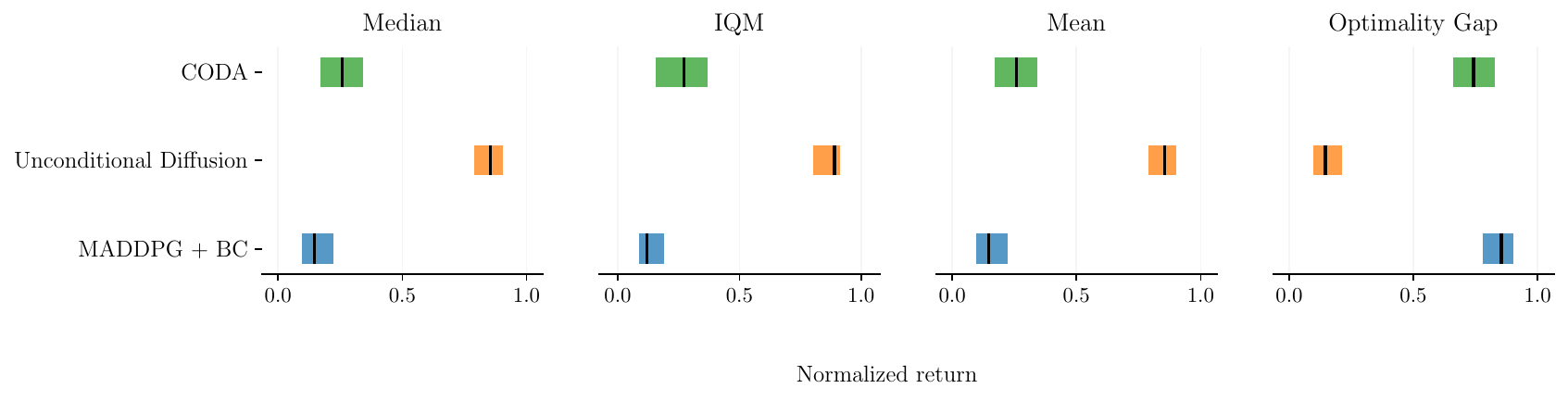}
    \caption{2HalfCheetah, Poor Dataset, BC 2.5, CODA guidance scale 0.6}
    \label{fig:halfcheetah-return_scores-poor}
\end{figure}

\begin{figure}[H]
    \centering
    \includegraphics[width=1\linewidth]{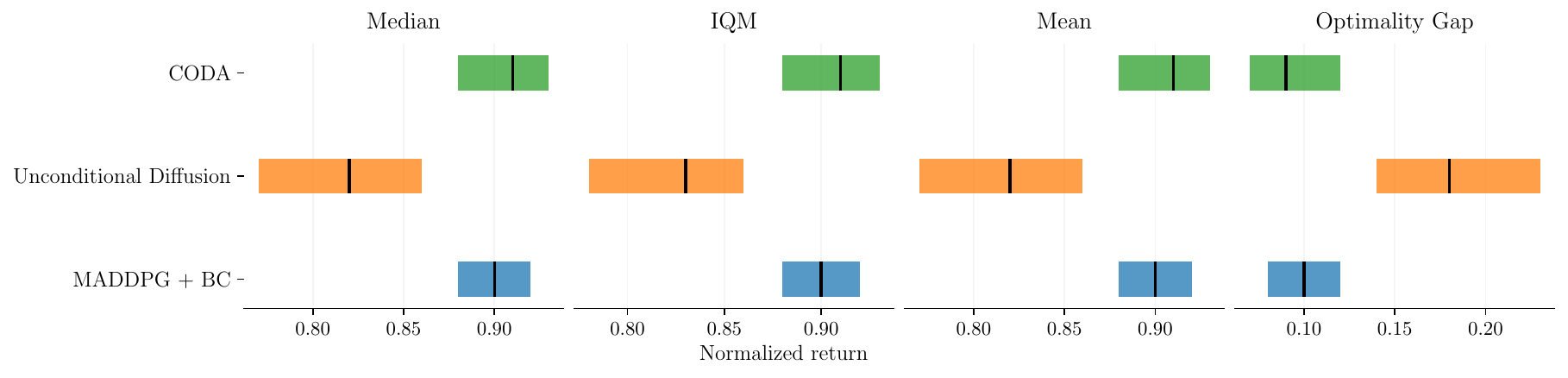}
    \caption{2HalfCheetah, Replay Dataset, BC 2.5, CODA guidance scale 0.6}
    \label{fig:halfcheetah-return_scores-replay}
\end{figure}

\begin{figure}[H]
    \centering
    \includegraphics[width=1\linewidth]{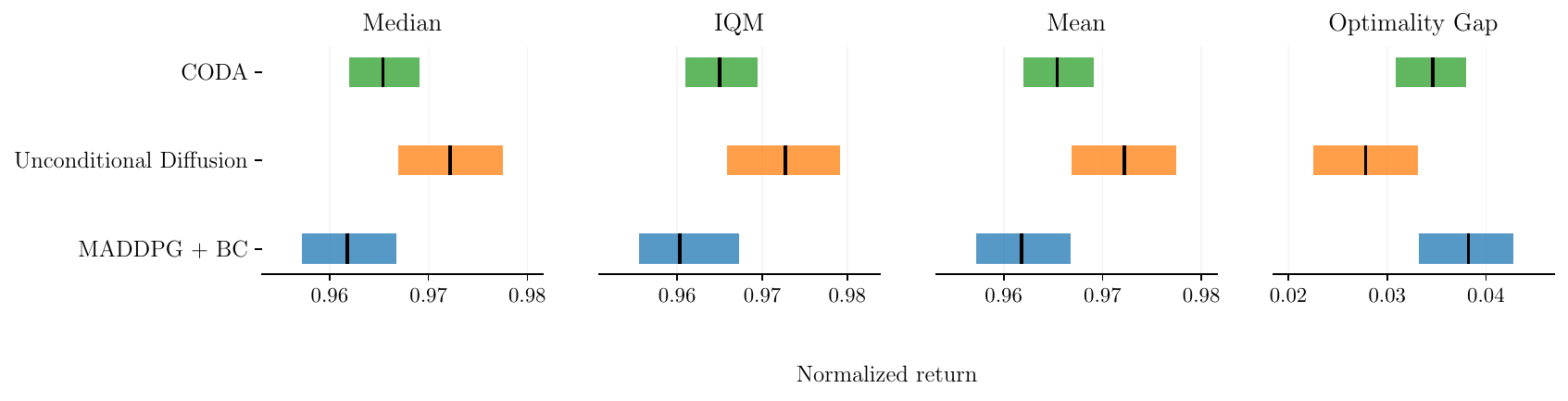}
    \caption{2HalfCheetah, Medium Dataset, BC 2.5, CODA guidance scale 0.6}
    \label{fig:halfcheetah-return_scores-medium}
\end{figure}

\begin{figure}[H]
    \centering
    \includegraphics[width=1\linewidth]{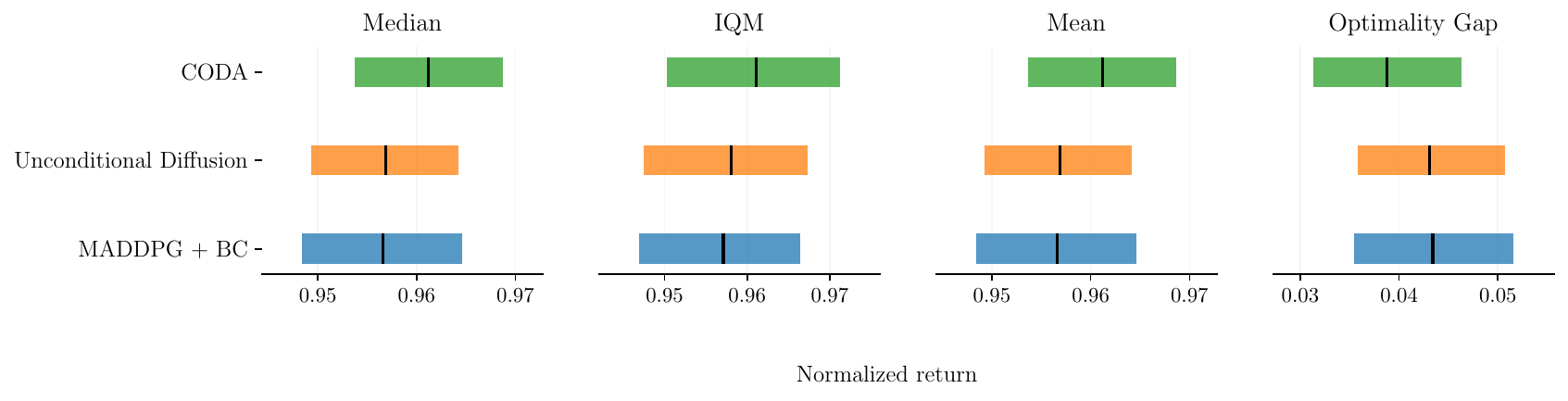}
    \caption{2HalfCheetah, Good Dataset, BC 2.5, CODA guidance scale 0.6}
    \label{fig:halfcheetah-return_scores-GOOD}
\end{figure}

\section{CODA Implementation Considerations}
\subsection{Policy Guided Diffusion Derivations}
\label{app:PGD_derviations}

\paragraph{Stabilization Techniques.} CODA inherits several stabilization techniques from prior work:
\begin{itemize}
    \item Cosine guidance schedules to modulate policy influence over the denoising trajectory \citep{Jackson2024PolicyGuidedD}.
    \item Score normalization to mitigate high-variance gradients.
    \item Applying guidance to denoised samples, rather than directly to noisy samples, following best practices in classifier-guided diffusion \citep{Guidance_diffusion_models}.
    \item Computing guidance gradients only with respect to the action components of the trajectory. \citet{Jackson2024PolicyGuidedD} observe that excluding state guidance improves stability. Intuitively, adapting actions while conditioning on plausible states is natural, whereas directly shifting states can move samples away from the data manifold of realistically encountered environment states. This can produce out-of-distribution states and lead to unstable guidance.
\end{itemize}

\subsection{Ensuring Diffusion Occurs in Unconstrained Space}
To enable stable and well-defined diffusion dynamics, we first map all constrained variables (e.g., bounded actions, observations, rewards, and terminal indicators) into an unconstrained real space before applying the noise perturbation and denoising processes. This transformation allows the diffusion model to operate over $\mathbb{R}^n$, where Gaussian noise assumptions hold naturally. If this step is omitted, samples tend to cluster or ``stick'' at the domain boundaries (e.g., actions saturating at their minimum or maximum values), since Gaussian noise can easily push constrained variables outside their valid range, leading to biased gradients and degenerate dynamics near the edges. By transforming into an unconstrained space, we ensure that diffusion proceeds in a smooth, consistent domain where additive Gaussian perturbations remain valid across all sample magnitudes.

\paragraph{Constrained to unconstrained transformation.}
For any variable $x$ bounded to $[\ell, h]$, we normalize it into the unit interval $[0, 1]$ and then apply the \emph{logit} transform:
\begin{equation}
    u = \mathrm{clip}\!\left(\frac{x - \ell}{h - \ell}, \varepsilon, 1 - \varepsilon\right), 
    \qquad
    z = \mathrm{logit}(u) = \log\!\frac{u}{1 - u},
\end{equation}
where $\varepsilon > 0$ is a small constant (e.g., $10^{-6}$) included to prevent numerical overflow at the boundaries $u \in \{0,1\}$, where the logit function diverges. This ensures finite gradients during optimization and avoids instability when values approach the domain limits. Note, this affine normalization maps each variable into the unit interval $[0,1]$ while preserving its empirical distribution shape (i.e., skewness and modality). It does not render the data uniform; it simply rescales the support to a standardized bounded domain. In practice, we actually apply a uniformization by applying an empirical CDF transform. This removes data skew and makes diffusion more efficient in general due to isotropic diffusion dynamics.
For unbounded quantities, we apply the identity mapping albeit still applying uniformization typically.
\paragraph{Diffusion in $\mathbb{R}^n$.}
After transformation, diffusion proceeds in the unconstrained space:
\begin{equation}
    z = z_0 + \sigma \, \epsilon, \qquad \epsilon \sim \mathcal{N}(0, I),
\end{equation}
where $z_0$ is the clean sample, $\sigma$ is the noise scale, and $\epsilon$ is Gaussian noise. The denoiser predicts $\hat{z}_0$ using the EDM preconditioning scheme, and the training loss is computed as a simple mean-squared error:
\begin{equation}
    \mathcal{L} = \| \hat{z}_0 - z_0 \|_2^2.
\end{equation}
No Jacobian correction is required since the diffusion process and loss are defined entirely in the transformed (unconstrained) space, not in the original variable domain. We do not compute any densities which would require a jacobian adjustment.
\paragraph{Inverse transformation.}
After sampling, the generated unconstrained outputs are mapped back to the original bounded domain using the sigmoid inverse:
\begin{equation}
    u = \sigma(z) = \frac{1}{1 + e^{-z}},
    \qquad
    x = \ell + (h - \ell) \, u.
\end{equation}
For terminal indicators (``dones''), $u$ can be interpreted as a probability in $[0,1]$ and optionally thresholded for binary execution \citep{10.5555/3666122.3668131}.
\paragraph{Summary.}
This two-way transformation (logit–sigmoid pair) ensures that the diffusion model operates in a mathematically unconstrained space, while respecting the physical or environmental bounds of the underlying variables. The inclusion of $\varepsilon$ terms guarantees numerical stability near the domain boundaries without altering the effective data distribution.

\subsection{Multiagent gradients}
The implication of \cref{eq:coda_guided_score} is that CODA derives guidance from a joint trajectory objective while applying gradients independently to each agent's action component. This preserves decentralized policy updates while still encouraging globally coordinated samples due to the coordination signal coming from the joint signal. 

\newpage
\section{Hyperparameters}
\subsection{Diffusion Models}
For the base diffusion model, we follow the EDM design of \citet{EDMKarras} and use a 1D temporal U-Net denoiser over full trajectory horizon. The denoiser operates on inputs of shape $[B,C,T]$, where $C$ is the joint trajectory feature dimension (concatenating the feature dimensions of $\tau=\Big(
s_0,\mathbf{o}_0,\mathbf{a}_0,r_0,\;
\dots,\;
s_H
\Big),$ and $H$ is the horizon. This concatenation enable multi-agent information sharing across the joint actions and joint observations. Noise-level embeddings, learned positional embeddings, and conditioning embeddings are fused in a shared latent space and injected into each encoder and decoder block through learned linear projections. 
\vspace{-100pt}
\begin{table}[H]
\centering
\caption{\textbf{Polynomial games: diffusion model and CFG configuration.} Diffusion architecture, conditioning setup, training noise distribution, and sampling schedule. The denoiser is a 1D temporal U-Net trained with classifier-free conditioning dropout and sampled with CFG guidance scale 1.0. Noise schedule and sigma parameterization are based on \citet{EDMKarras}. Learning rate was ablated over the range $1 \times 10^{-1}$ to $1 \times 10^{-3}$.}
\begin{tabular}{ll}
\hline
\textbf{Diffusion architecture} & \textbf{Value} \\
\hline
Denoiser backbone & 1D temporal U-Net \\
U-Net embedding & 32 \\
U-Net blocks & 2 \\
Kernel size & 3 \\
Trajectory horizon $H$ & 1 \\
Positional embeddings & Learned \\
Condition embedding dimension & 32 \\
Condition dropout & 0.1 \\
CFG guidance scale & 1.0 \\
\hline
\textbf{Noise schedule (sampling)} &  \\
\hline
Schedule type & Karras (EDM) \\
Minimum sampling noise $\sigma_{\min}$ & 0.002 \\
Maximum sampling noise $\sigma_{\max}$ & 80.0 \\
Schedule curvature $\rho$ & 7.0 \\
Diffusion steps & 40 \\
\hline
\textbf{Noise distribution (training)} &  \\
\hline
Distribution & Log-normal over $\sigma$ \\
Log-mean of training noise $\mu_{\log \sigma}$ & $-1.2$ \\
Log-std of training noise $\sigma_{\log \sigma}$ & $1.2$ \\
Training noise range & $[\sigma_{\min}, \sigma_{\max}]$ \\
\hline
\textbf{Training} &  \\
\hline
Batch size & 1000 \\
Learning rate & $1\times10^{-2}$ \\
Training epochs & 10000 \\
Normalizer & CDF Normalizer \\
\hline
\textbf{Sampling} &  \\
\hline
Generated samples & 1000 \\
\hline
\end{tabular}
\label{tab:diffusion_hparams}
\end{table}
\begin{table}[t]
\centering
\caption{\textbf{MaMuJoCo: Diffusion model and Classifier-based configuration.} The denoiser is a 1D temporal U-Net. Noise schedule and sigma parameterization are based on \citet{EDMKarras}. Learning rate was ablated over the range $1 \times 10^{-1}$ to $1 \times 10^{-3}$. Guidance scale is ablated over range $1 \times 10^{-3}$ to $1 \times 10^{1}$ to find optimal performance in each environment.}
\begin{tabular}{ll}
\hline
\textbf{Diffusion architecture} & \textbf{Value} \\
\hline
Denoiser backbone & 1D temporal U-Net \\
U-Net embedding & 16 \\
U-Net blocks & 2 \\
Kernel size & 3 \\
Trajectory horizon $H$ & 20 \\
Positional embeddings & Learned \\
Condition embedding dimension & 16 \\
Classifier guidance scale & [0.6, 1.0] \\
\hline
\textbf{Noise schedule (sampling)} &  \\
\hline
Schedule type & Karras (EDM) \\
Minimum sampling noise $\sigma_{\min}$ & 0.002 \\
Maximum sampling noise $\sigma_{\max}$ & 80.0 \\
Schedule curvature $\rho$ & 7.0 \\
Diffusion steps & 40 \\
\hline
\textbf{Noise distribution (training)} &  \\
\hline
Distribution & Log-normal over $\sigma$ \\
Log-mean of training noise $\mu_{\log \sigma}$ & $-1.2$ \\
Log-std of training noise $\sigma_{\log \sigma}$ & $1.2$ \\
Training noise range & $[\sigma_{\min}, \sigma_{\max}]$ \\
\hline
\textbf{Training} &  \\
\hline
Batch size & 1000 \\
Learning rate & $1\times10^{-3}$ \\
Training epochs & 10000 \\
Normalizer & CDF Normalizer \\
\hline
\textbf{Sampling} &  \\
\hline
Generated samples & 1000 \\
\hline
\end{tabular}
\label{tab:diffusion_hparams}
\end{table}

\subsection{MADDPG}
\begin{table}[H]
\centering
\caption{\textbf{MADDPG configuration for polynomial games.} 
In the stateless environment setting, each agent's policy is parameterized as a single trainable action bias independent of observations. Learning rate ablated over $1 \times 10^{-3}$ to $1 \times 10^{-1}$}
\begin{tabular}{ll}
\hline
\textbf{MADDPG Agent} & \textbf{Value} \\
\hline
Agent type & Stateless deterministic policy \\
Number of agents & 2 \\
Observation dependence & None (stateless environment) \\
Action dimension & 1 per agent \\
Actor parameterization & Trainable bias per agent \\
Policy initialization & $[-0.64,\;0.65]$ \\
Action bounds & $[-1, 1]$ \\
\hline
\textbf{Critic (centralized)} &  \\
\hline
Critic usage & Not used for policy learning in stateless setting \\
\hline
\textbf{Training} &  \\
\hline
Optimizer & SGD \\
Learning rate & $1 \times 10^{-1}$ \\
Gradient clipping & 1.0 \\
Training batch size & 64 \\
Burn-in iterations & 2 \\
Number of transitions sampled & 4000  \\
\hline
\end{tabular}
\label{tab:maddpg_config}
\end{table}

\newpage
For the MuJoCo experiments, we use MADDPG augmented with a behaviour cloning loss \citep{ogmarl}. Each agent is parameterized by a recurrent policy network operating on local observations, while training uses centralized state--joint-action critics and soft target network updates.
\begin{table}[H]
\centering
\caption{\textbf{MaMuJoCo experiments: MADDPG + BC configuration.} Hyperparameters and network architecture used for the MuJoCo setting. Policies are parameterized by recurrent networks operating on local observations, while critics are centralized state--joint-action value networks. Behaviour cloning is incorporated with coefficient $\alpha_{\mathrm{BC}}=2.5$. This coefficient was ablated over the range $0$ to $2.5$.}
\begin{tabular}{ll}
\hline
\textbf{MuJoCo MADDPG + BC} & \textbf{Value} \\
\hline
Discount factor $\gamma$ & 0.99 \\
Target update rate $\tau$ & 0.005 \\
Critic learning rate & $3 \times 10^{-4}$ \\
Policy learning rate & $3 \times 10^{-4}$ \\
BC coefficient $\alpha_{\mathrm{BC}}$ & 2.5 \\
Add agent ID to observation & True \\
\hline
\textbf{Policy network} &  \\
\hline
Policy type & Recurrent policy network \\
Input & Local observations \\
Linear layer width & 64 \\
Recurrent hidden dimension & 64 \\
Output & Agent action \\
\hline
\textbf{Critic network} &  \\
\hline
Critic type & Centralized state--joint-action critic \\
Number of critics & 2 \\
Critic inputs & State and joint action \\
\hline
\end{tabular}
\label{tab:mujoco_maddpg_bc}
\end{table}

%% file: 3_algorithms.tex
\begin{algorithm}[H]
\caption{CODA: Trajectory Sampling via Joint Policy–Guided Diffusion (Multi-Agent)}
\label{alg:coda-sampling}
\begin{algorithmic}[1]
\State \textbf{Parameters:} noise schedule $\{\sigma_n\}_{n=0}^{N_{\text{diff}}}$, guidance schedule $\{\lambda_n\}_{n=0}^{N_{\text{diff}}}$, temporary noise factor $\gamma_n$, noise level $S_{\text{noise}}$, diffusion steps $N_{\text{diff}}$
\State \textbf{Require:} trajectory denoiser $D_\theta$ trained on offline data; joint target policy $\boldsymbol{\pi}_\phi = \{\pi^i_\phi\}_{i=1}^N$
\State \textbf{Notation:} a trajectory $\tau = (s_0,\mathbf{o}_0 ,\mathbf{a}_0,r_0,\ldots,s_{H})$; bar $\bar{\cdot}$ denotes denoised estimates; hat $\hat{\cdot}$ noised variables
\State \textbf{Init:} sample $\tau_0 \sim \mathcal{N}(0,\sigma_0^2 I)$  \Comment{EDM/score-based ODE init}
\For{$n=0$ \textbf{to} $N_{\text{diff}}-1$}
  \State sample $\epsilon_n \sim \mathcal{N}(0,S_{\text{noise}}^2 I)$
  \State $\hat{\sigma}_n \gets \sigma_n + \gamma_n \sigma_n$; \quad $\hat{\tau}_n \gets \tau_n + \sqrt{\hat{\sigma}_n^2-\sigma_n^2}\,\epsilon_n$
  \State $\bar{\tau}_n \gets D_\theta(\hat{\tau}_n;\hat{\sigma}_n)$ \Comment{Denoised trajectory, converted to constrained space}
  \State $d_n \gets (\hat{\tau}_n - \bar{\tau}_n)/\hat{\sigma}_n$ \Comment{EDM preconditioning: $\partial \tau/\partial \sigma$ Score approx}
  \State \textbf{// Joint policy guidance on actions only (multi-agent)}
  \State Extract $(\bar{\mathbf{a}}_t,\bar{\mathbf{o}}_t)$ for all $t$ from $\bar{\tau}_n$
  \For{each time $t=0,\ldots,H-1$}
     \State $g_{t} \gets \sum_{i=1}^{N} \nabla_{\bar{a}^{(i)}_t} \log \pi^{(i)}_\phi\!\left(\bar{a}^{(i)}_t \mid \bar{o}^{(i)}\right)$ \Comment{Per-agent scores}
  \EndFor
  \State $g_n \gets \text{stack}(g_t)_{t=0}^{H-1}$; \quad $g_n \gets g_n / \|g_n\|_2$ \Comment{Normalize for stability}
  \State $\hat{\mathbf{a}}_n \gets \text{actions}(\hat{\tau}_n)$; \quad $\hat{\mathbf{a}}_n \gets \hat{\mathbf{a}}_n + \lambda_n\, g_n$ \Comment{Apply guidance to \emph{noised} actions}
  \State $\hat{\tau}_n \gets \text{replace\_actions}(\hat{\tau}_n,\hat{\mathbf{a}}_n)$
  \State $\tau_{n+1} \gets \hat{\tau}_n + (\sigma_{n+1}-\hat{\sigma}_n)\, d_n$ \Comment{Euler step}
  \If{$\sigma_{n+1} \neq 0$} \Comment{2nd-order correction (EDM)}
     \State $d'_n \gets \left(\tau_{n+1} - D_\theta(\tau_{n+1};\sigma_{n+1})\right)/\sigma_{n+1}$ \Comment{Score approx}
     \State $\hat{\tau}_{n+1} \gets \hat{\tau}_n + (\sigma_{n+1}-\hat{\sigma}_n)\,\big(\tfrac{1}{2}d_n + \tfrac{1}{2}d'_n\big)$
     \State $\tau_{n+1} \gets \hat{\tau}_{n+1}$
  \EndIf
\EndFor
\State \textbf{return} $\tau_{N_{\text{diff}}}$ \Comment{Map back to original domains if using logit/sigmoid transforms}
\end{algorithmic}
\end{algorithm}

\begin{algorithm}[H]
\caption{Training an Agent with CODA}
\label{alg:coda-training}
\begin{algorithmic}[1]
\State \textbf{Parameters:} epochs $N_{\text{epochs}}$, policy update steps per epoch $N_{\text{policy}}$, synthetic set size $B$ (trajectories), guidance schedule $\{\lambda_n\}_{n=0}^{N_{\text{diff}}}$, diffusion steps $N_{\text{diff}}$, mix ratio $\alpha \in [0,1]$
\State \textbf{Require:} offline dataset $\mathcal{D}_{\text{off}}$; diffusion sampler $\mathcal{F} \equiv \text{\cref{alg:coda-sampling}}$; base offline RL algorithm (e.g., MADDPG-BC)
\State Train trajectory diffusion $D_\theta$ on $\mathcal{D}_{\text{off}}$ \Comment{Behavior prior over full trajectories}
\State Initialize joint policy $\boldsymbol{\pi}_\phi$ (and critics if applicable)
\For{$e=1$ \textbf{to} $N_{\text{epochs}}$}
   \State \textbf{// Generate on-policy synthetic experience under current joint policy}
   \State $\mathcal{D}_{\text{syn}} \gets \{\tau^{(j)} \sim \mathcal{F}(\,\cdot\,|\boldsymbol{\pi}_\phi, D_\theta;\{\lambda_n\}_{n=0}^{N_{\text{diff}}})\}_{j=1}^{B}$
   \State \textbf{// Replay mix with real data}
   \State $\mathcal{B} \gets \text{concat}\big(\text{sample}(\mathcal{D}_{\text{syn}},\alpha),\ \text{sample}(\mathcal{D}_{\text{off}},1-\alpha)\big)$
   \For{$k=1$ \textbf{to} $N_{\text{policy}}$}
      \State Sample a mini-batch of sub-trajectories/transitions from $\mathcal{B}$ (or $\mathcal{D}_{\text{syn}}$ if $\alpha{=}1$)
      \State Update critics and each agent policy $\pi_\phi^{(i)}$ via chosen offline RL algo   ; 
   \EndFor
\EndFor
\State \textbf{return} $\boldsymbol{\pi}_\phi$
\end{algorithmic}
\end{algorithm}